\documentclass[sigconf]{acmart}
\AtBeginDocument{%
  }
\setcopyright{acmlicensed}
\copyrightyear{2025}
\acmYear{2025}
\setcopyright{cc}
\setcctype{by}
\acmDOI{10.1145/3746027.3754837}
\acmConference[MM '25]{Proceedings of the 33rd ACM International Conference on Multimedia}{October 27--31, 2025}{Dublin, Ireland}
\acmBooktitle{Proceedings of the 33rd ACM International Conference on Multimedia (MM '25), October 27--31, 2025, Dublin, Ireland}
\acmISBN{979-8-4007-2035-2/2025/10}

\usepackage[utf8]{inputenc}
\usepackage{graphicx}
\usepackage{booktabs}
\usepackage{makecell}
\usepackage{multirow}
\usepackage{xspace}
\usepackage{soul}
\usepackage{xcolor}
\usepackage{array}
\usepackage{pifont}    
\usepackage[ruled,vlined,linesnumbered]{algorithm2e}
\usepackage{amsmath}
\usepackage{enumitem}      
\usepackage{arydshln}
\usepackage[normalem]{ulem}
\useunder{\uline}{\ul}{}

\newcommand{\benchmark}{\textit{D-Judge}\xspace}
\newcommand{\data}{\textit{D-ANI}\xspace}
\newcommand{\codeurl}{\url{https://github.com/ryliu68/DJudge}}
\newcommand{\dataurl}{\url{https://huggingface.co/datasets/Renyang/DANI}}

\usepackage[font=small,skip=4pt]{caption} 
\usepackage{caption}
\begin{document}

\title[\benchmark]{\benchmark: How Far Are We? Assessing the Discrepancies Between AI-synthesized and Natural Images through Multimodal Guidance}


\author{Renyang Liu}
\authornote{This work was conducted while Renyang Liu was a Research Intern at the School of Cyber Science and Technology, Shenzhen Campus of Sun Yat-sen University.}
\affiliation{%
  \institution{
       School of Cyber Science and Technology, Shenzhen Campus of Sun Yat-sen University
  }
  \city{Shenzhen}
  \state{Guangdong}
  \country{China}
}
\affiliation{%
  \institution{Institute of Data Science,
  National University of Singapore}
  \city{Singapore}
  \country{Singapore}
  }
\email{ryliu@nus.edu.sg}

\author{Ziyu Lyu}
\authornote{Ziyu Lyu is the corresponding author.}
\affiliation{%
  \institution{
       School of Cyber Science and Technology, Shenzhen Campus of Sun Yat-sen University
  }
  \city{Shenzhen}
  \state{Guangdong}
  \country{China}
}
\email{lvzy7@mail.sysu.edu.cn}

\author{Wei Zhou}
\affiliation{%
  \institution{College of Modern Engineering and the Engineering Research Center of Cyberspace, Yunnan University}
  \city{Kunming}
  \state{Yunnan}
  \country{China}
  }
\email{zwei@ynu.edu.cn}

\author{See-Kiong Ng}
\affiliation{%
  \institution{
    Institute of Data Science, 
    National University of Singapore}
  \city{Singapore}
  \country{Singapore}
  }
\email{seekiong@nus.edu.sg}

\renewcommand{\shortauthors}{Renyang Liu et al.}

\begin{abstract}
In the rapidly evolving field of Artificial Intelligence Generated Content (AIGC), a central challenge is distinguishing AI-synthesized images from natural ones. Despite the impressive capabilities of advanced generative models in producing visually compelling images, significant discrepancies remain when compared to natural images. To systematically investigate and quantify these differences, we construct a large-scale multimodal dataset, \data, comprising 5,000 natural images and over 440,000 AIGI samples generated by nine representative models using both unimodal and multimodal prompts, including Text-to-Image (T2I), Image-to-Image (I2I), and Text-and-Image-to-Image (TI2I). We then introduce an AI–Natural Image Discrepancy assessment benchmark (\benchmark) to address the critical question: \textit{how far are AI-generated images (AIGIs) from truly realistic images?} Our fine-grained evaluation framework assesses the \data dataset across five dimensions: naive visual quality, semantic alignment, aesthetic appeal, downstream task applicability, and coordinated human validation. Extensive experiments reveal substantial discrepancies across these dimensions, highlighting the importance of aligning quantitative metrics with human judgment to achieve a comprehensive understanding of AI-generated image quality. The code\footnote{\textcolor{blue}{\codeurl}} and data\footnote{\textcolor{blue}{\dataurl}} are publicly available.
\end{abstract}



\maketitle

\section{Introduction}
\label{sec:intro}
With the rapid advancement of deep learning techniques, the proliferation of Artificial Intelligence Generated Content (AIGC) has garnered significant attention across various domains, such as e-commerce, gaming, medicine, animation, and autonomous driving \cite{li2024distrifusion,qian2024boosting}.  
AI-generated Images (AIGI) have been one of the mainstream forms of AIGC, and a variety of AI image generative models have been proposed to make the generated synthetic images as realistic as natural images, ranging from the earlier generative adversarial networks (GANs) \cite{cvpr/TaoB0X23,cvpr/Tao00JBX22}, advanced diffusion models (DMs) \cite{iccv/XuWZWS23,iccv/WeiZJB0Z23} to the large multimodal generative model like DALL·E \cite{corr/abs-2204-06125}.

Despite the rapid development of AI image generative models, AI-generated images still fall short of real-world application standards due to discrepancies with realistic natural images \cite{tcsvt/AGIQA-3K}. To address this, various AI-generated image quality assessment methods have emerged \cite{corr/PKU-I2IQA,nips_23/ImageReward,iccv/HuLKWOKS23}. For instance, Wang et al. \cite{corr/PKU-I2IQA} introduced the AIGCIQA2023 database and a subjective evaluation framework based on quality, authenticity, and text-image correspondence. Similarly, PKU-I2IQA \cite{corr/PKU-I2IQA} developed a perception-based database for image-to-image generation using no-reference and full-reference methods. Other studies, such as Pick-a-Pic \cite{nips/KirstainPSMPL23}, HPS v2 \cite{corr/abs-2306-09341}, TIFA \cite{iccv/HuLKWOKS23}, and ImageReward \cite{nips_23/ImageReward}, focused on training unified score models for automatic AIGC image quality assessment, moving beyond traditional metrics like Inception Score \cite{nips/SalimansGZCRCC16} and FID \cite{nips/FID}.

Despite advancements in AIGC research, several critical issues remain unresolved. \textbf{First}, there is a lack of systematic and comprehensive studies addressing the discrepancies between AI-generated images and realistic natural images, which is crucial for enabling the practical application of AIGC images in real-world scenarios and achieving meaningful breakthroughs in the field. Most prior studies primarily focus on perceptual quality using traditional image quality metrics, text-to-image correspondence, or subjective human preferences, leaving a gap in understanding the broader implications of these discrepancies.  
\textbf{Second}, previous research predominantly considers unimodal prompts, such as T2I or I2I, while neglecting multimodal prompts like TI2I, and the datasets used are often small, typically comprising only a few thousand images, which limits their capacity for comprehensive analysis and evaluation.

\begin{figure}[t]
    \centering
    \includegraphics[width=0.45\textwidth]{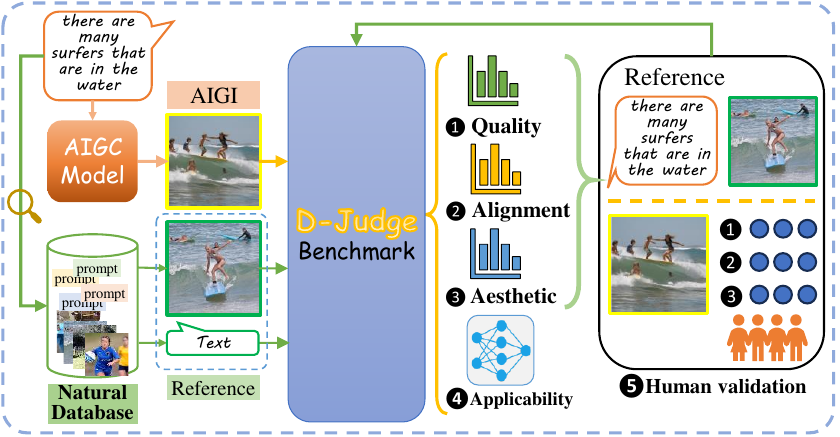}
    \caption{Toy example demonstrating the key aspects of \benchmark.}
    \label{fig:toy_example}
    \setlength{\abovecaptionskip}{5pt}    
    \vspace{-10pt}
\end{figure}

To address the above challenges, we proposed an AI-Natural Image Discrepancy Evaluation Benchmark (\benchmark) to investigate and interpret what discrepancies still remained between AI-generated images and natural images, finally answering the important question “How far are AI generative models with respect to the visual forms of images?”. Especially, we have two important contributions with reference to the key problems. \textbf{Large Multimodal Evaluation Dataset construction:} we have curated a large multimodal evaluation dataset called \textbf{\data} (Distinguishing Natural and AI-generated) dataset, with about 440,000 AIGI generated from 9 representative generative models based on both unimodal and mutimodal guidance including Text-to-Image (T2I), Image-to-Image (I2I), and Text-and-Image-to-Image (TI2I). The data size of our DNAI dataset scales to 100x that of prior datasets. \textbf{Fine-grained Evaluation Framework}, we propose a fine-grained evaluation framework to conduct the systematic and comprehensive assessment and evaluation of DNAI, covering 5 diverse and important aspects, including naive visual feature quality, semantic alignment among multimodal generation, aesthetic appeal, downstream applicability, and coordinated human validation. The toy example is illustrated in Figure~\ref{fig:toy_example}. For example, given a text prompt, an image is generated using a generative model and paired with a natural reference image retrieved from the natural database. \benchmark then evaluates the discrepancy rate between the two images across the aforementioned dimensions.

Through our \data dataset and fine-grained evaluation framework \benchmark, we conduct extensive benchmark analysis and evaluation and conclude key insights to answer the discrepancy questions:
\begin{itemize}[nosep, leftmargin=*]
    \item \textbf{Significant discrepancies in key Areas:} \textit{AIGIs exhibit substantial discrepancies from natural images in terms of quantitative measures from all aspects, up to 40\%, 33.57\%, 27.66\%, and 96.95\% in frame-level image quality, semantic alignment, aesthetic appeal, and downstream applicability, respectively.} 
    \item \textbf{Multimodal Alignment:} Different prompted generations might have different semantic alignment discrepancies compared to natural ones. \textit{Generated images prompted with texts including both T2I and TI2I demonstrated less semantic discrepancy than I2I.}    
    \item \textbf{Downstream Task Applicability:} \textit{Significant differences in usability between AI-generated and natural images are observed in downstream tasks, particularly in fine-grained object recognition and VQA, with discrepancy rates reaching as high as 94.29\% and 96.95\%, respectively. This underscores the need for further advancements in generative models to enhance their practical applicability in real-world scenarios.}
    \item \textbf{Human Evaluation \textit{vs.} Quantitative Metrics:} \textit{Human evaluation results reveal larger discrepancies compared with quantitative ones.} It validates the necessity of incorporating human evaluation for coordination.
\end{itemize}

\section{Related Work}
\label{sec:related}

\paragraph{Natural Image Evaluation:} Over the past decades, numerous image quality assessment methods have been developed to evaluate natural images \cite{ncc/NDBCM15_piqe}. These methods have focused on various visual features and properties, such as perceptual appearance \cite{cvpr/LPIPS}, naturalness \cite{tip/MaLZDWZ18_liqe}, and aesthetics \cite{tip/TalebiM18_nima}. For instance, BRISQUE evaluates natural image quality by analyzing spatial domain features, while PIQE assesses perceptual quality through block-based image segmentation, both serving as effective no-reference metrics. Other notable measures include FID and Inception Scores, which assess the quality and diversity of generated images, and SSIM \cite{tip/WangBSS04} and PSNR, which quantify structural similarity and structure-level accuracy. Beyond quantitative measures, human evaluations have also been leveraged to estimate natural image quality. For example, NIMA \cite{tip/TalebiM18_nima}, proposed by Talebi and Milanfar, uses a deep CNN trained on human-rated images to predict aesthetic quality. Similarly, Wong et al. introduced the AVA dataset, which facilitates aesthetic assessment based on human preferences \cite{cvpr/MurrayMP12}.

\paragraph{AI-generated Image Evaluation:} Recently, AI-generated content (AIGC) has seen significant advancements with the rise of generative models. Several methods and benchmarks have been introduced to evaluate AI-generated images \cite{icmcs/AGIQA-1K,iccv/HuLKWOKS23}. These methods primarily focus on three aspects: perceptual quality, text-image correspondence, and aesthetics \cite{nips_23/ImageReward,corr/PKU-I2IQA,icmcs/QRefine,cvpr_25/LiTLZDWJLMLLZ25,tcsv_25/ChenSWZJHMZZ25}, often training unified models to assess the overall quality of AI-generated images. For example, ImageReward \cite{nips_23/ImageReward} presents a framework for aligning image generation with human preferences, AGIQA-3k \cite{tcsvt/AGIQA-3K} offers a comprehensive benchmark for assessing AI-generated image quality, and QBench \cite{iclr/0001Z0CLWLSYZL24} evaluates content quality across multiple dimensions.

Despite advancements, little research has delved into the fine-grained differences between AI-generated and natural images, and existing datasets are often too small for comprehensive evaluation. To address these gaps, we propose D-Judge as AI-Natural image difference Assessment Benchmark, featuring a large Distinguishing AI-Natural Image Dataset and a systematic, fine-grained evaluation framework to thoroughly explore these differences.

\section{\benchmark}
\label{sec:framework}

\begin{figure}[t]
    \centering
    \includegraphics[width=0.45\textwidth]{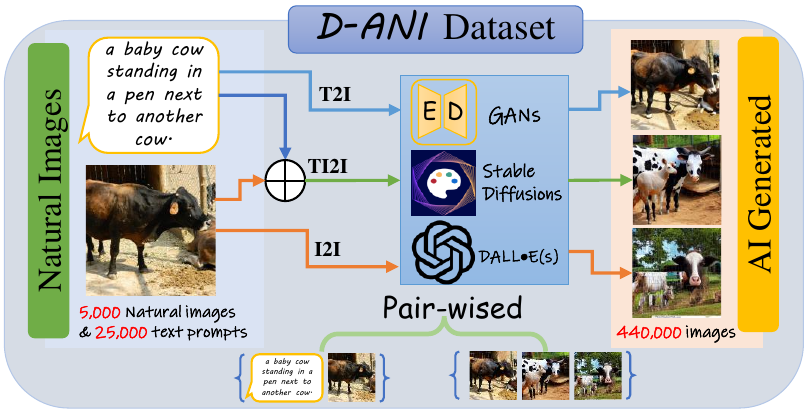}
    \caption{The \data dataset of \benchmark, guided by three types of prompts. Each generated image is paired with both the corresponding text and image from the naturally collected dataset (e.g., COCO).}
    \label{fig:dataset}
\end{figure}

\begin{figure*}[ht]
    \centering
    \includegraphics[width=\textwidth]{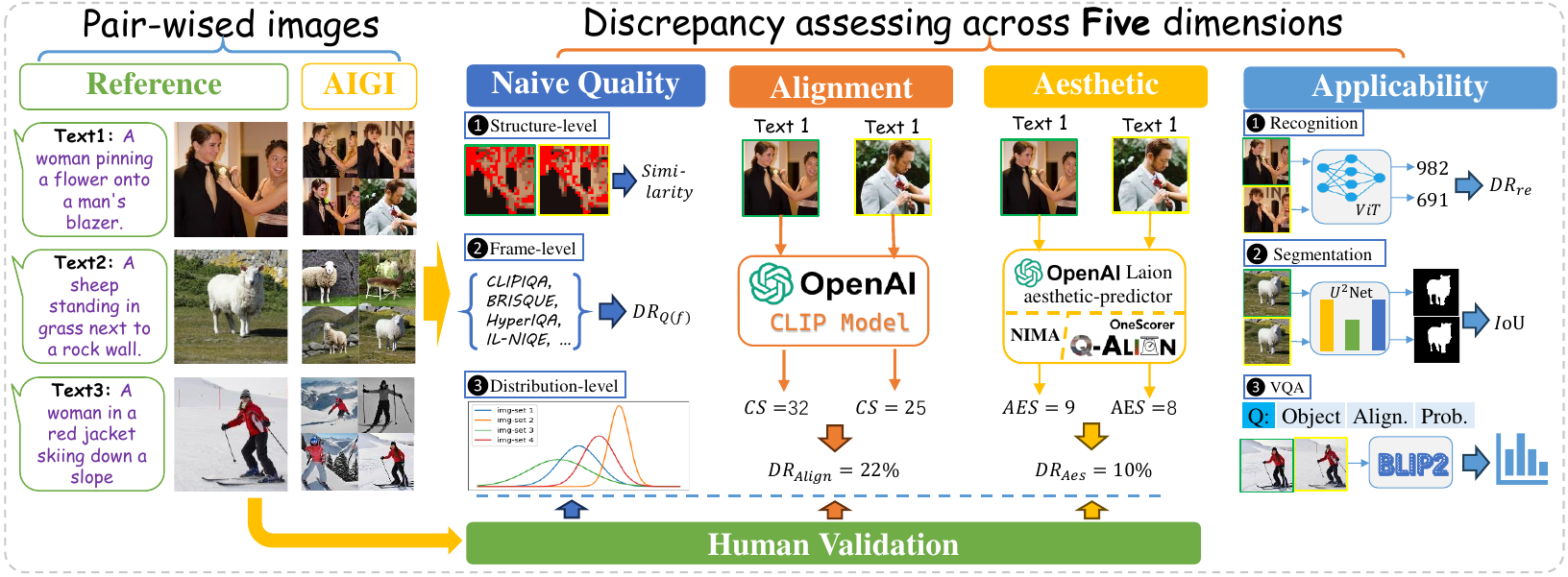}
    \caption{The framework of \benchmark evaluates the \textbf{Discrepancy Rate ($DR$)} between AI-generated images and natural reference images across \textbf{five} dimensions: \ding{202} \textit{Naive Quality}, \ding{203} \textit{Alignment}, \ding{204} \textit{Aesthetic}, and \ding{205} \textit{Applicability}, complemented by \ding{206} \textbf{human validation}. The images with \textbf{green borders} represent reference natural images, while those with \textbf{yellow borders} are AI-generated counterparts. This comprehensive evaluation integrates structure-level, frame-level, and distribution-level quality checks, alongside advanced tools like \textit{OpenAI CLIP}, \textit{OpenAI Laion Aesthetic Evaluator}, \textit{QAlign}, \textit{ViT}, \textit{U\textsuperscript{2}Net}, and \textit{BLIP2} to robustly assess the gap between the two image types. Where $CS$ means CLIP Score for the text and given images, and $AES$ means Aesthetic score.}

    \label{fig:framework}
\end{figure*}

To tackle the challenges of current AI-generated image evaluation, we construct an AI-Natural Image Discrepancy evaluation benchmark (\textbf{\benchmark}) to assess the potential discrepancies between AI-generated images and natural images and answer how far AIGIs are from Natural Images. Firstly, a Distinguishing AI-Natural Image (\data) Dataset is constructed, in which we collect AI-generated images from diverse generative models for real natural images in MS COCO dataset \cite{eccv/COCO} based on three types of guidance prompts, e.g., text-only, image-only and text-image prompts, as shown in Figure~\ref{fig:dataset}; Then, we devise a systematic and comprehensive evaluation framework as Figure~\ref{fig:framework} illustrated to measure and evaluate the potential differences between AI-generated images and real natural images from five aspects including naive image quality, semantic alignment, aesthetic appeal, downstream applicability, and the coordinated human assessment. In the following parts, we introduce the \data Dataset and the evaluation framework in detail.

\subsection{\data Dataset}

\subsubsection{Natural Images}
We constructed the Distinguished Natural and AI-generated Image dataset (\data) based on the classical MS COCO natural image dataset \cite{eccv/COCO}. Specifically, we selected 5,000 carefully curated images from the MS COCO validation set, each accompanied by at least five unique captions, resulting in a total of 25,000 distinct text-image pairs. This rich textual diversity enhances the dataset's applicability across various image-to-text and text-to-image tasks, making it an invaluable resource for evaluating and training image generation and understanding models. The 25,000 text-image pairs form the \textbf{referenced natural image set}, guiding generative models in producing new images in our study.

\subsubsection{Generated Images}
We collected AI-generated images from nine representative generative models using three distinct types of prompts: text-only (Text-to-Image, T2I), image-only (Image-to-Image, I2I), and combined text-image (Text-and-Image-to-Image, TI2I). These models range from earlier generative adversarial networks (GANs) like GALIP \cite{cvpr/TaoB0X23} and DF-GAN \cite{cvpr/Tao00JBX22}, to recent diffusion models (DMs) such as Stable Diffusion (versions v1.4, v1.5, v2.1, and XL) \cite{cvpr/RombachBLEO22}, Versatile Diffusion (VD) \cite{iccv/XuWZWS23}, and OpenAI's commercial models DALL·E 2 (D·E 2) \cite{corr/abs-2204-06125} and DALL·E 3 (D·E 3) \cite{dalle3}, which generate images via OpenAI's API, while other models generate images locally. Detailed information \textit{w.r.t} these models can be found in Appendix~\ref{sec_app:dataset}. 

Each model generated 25,000 images per guidance mode, except for DALL·E 2 and DALL·E 3, which produced 5,000 images per mode, leading to a total of 440,000 images. Due to intrinsic guidance capabilities, DF-GAN, GALIP, and DALL·E 3 were exclusively used in T2I mode, whereas DALL·E 2 operated in both T2I and I2I modes. The details are shown in Table~\ref{tab:DANI_dataset}. 

The image resolution varies across these models: GALIP produces images at 224x224 pixels, DF-GAN at 256x256 pixels, models such as various versions of Stable Diffusion and DALL·E 2 at 512x512 pixels, and DALL·E 3 provides the highest resolution of 1024x1024 pixels. These generated images, combined with the referenced natural images, constitute our distinguishing AI-Natural image dataset. 

\begin{table}[t]
    \centering
    \small
    \caption{The details of the \data dataset.}
    \label{tab:DANI_dataset}
    \renewcommand{\arraystretch}{1.2}
    \setlength\tabcolsep{2pt}
    \resizebox{1\columnwidth}{!}{
    \begin{tabular}{cccccccccc}
    \toprule
    Types & DFGAN & GALIP & SD14 & SD15 & VD & SD21 & SDxl & D·E 2 & D·E 3 \\ \midrule
    T2I   & 25,000 & 25,000 & 25,000   & 25,000   & 25,000      & 25,000   & 25,000 & 5,000 & 5,000   \\
    I2I   & -     & -     & 25,000   & 25,000   & 25,000      & 25,000   & 25,000 & 5,000 & -    \\
    TI2I  & -     & -     & 25,000   & 25,000   & 25,000      & 25,000   & 25,000 & -  & -     \\ \bottomrule
    \end{tabular}
    }
\end{table}

\begin{table}[ht]
    \centering
    \small
    \caption{Comparison of \data dataset and Counterparts.}
    
    \label{tab:aigc_dataset}
        \renewcommand{\arraystretch}{1.2}
    \setlength\tabcolsep{2.4pt}
    \begin{tabular}{lccccc}
    \toprule
    Dataset       & Model                             & Text                                      & Image                                    & Text \textit{vs.} Image                             & AIGI                                       \\ \midrule
    AGIQA-1K~\cite{icmcs/AGIQA-1K}      & 2                                     & 1,080                                      & -                                        & -                                         & 1,080                                       \\
    AGIQA-3K~\cite{tcsvt/AGIQA-3K}      & 6                                     & 300                                       & -                                        & -                                         & 2,982                                       \\
    AGIQA-20K~\cite{cvprW/AIGIQA-20K}      & \textbf{15}                                     & \textbf{200,000}                                       & -                                        & -                                         & 200,000                                       \\
    AIGCIQA2023~\cite{cicba/AIGCIQA2023}   & 6                                     & 100                                       & -                                        & -                                         & 2,400                                       \\
    ImageReward~\cite{nips_23/ImageReward}     & 3                                     & 8,878                                       & -                                      & -                                         & 136,892                                       \\
    HPDv2~\cite{corr/HPDv2}     & 9                                     & 108,000                                       & -                                      & -                                         & 381,304                                      \\

    PKU-I2IQA~\cite{corr/PKU-I2IQA}     & 2                                     & 200                                       & 200                                      & -                                         & 1,600                                       \\
    
    \textbf{\data} (Ours) & 9 & 25,000 & \textbf{5,000} & \textbf{25,000} & \textbf{440,000} \\ \bottomrule
    \end{tabular}
        
    \vspace{-1em}
\end{table}

Table~\ref{tab:aigc_dataset} presents the key statistics and characteristics of our proposed \data dataset in comparison with existing AIGC evaluation datasets. Specifically, the \data dataset consists of 440,000 AI-generated images produced by nine representative generative models, alongside 25,000 human-annotated image-text pairs from the MS COCO validation set. This results in a total of \textbf{445,000} high-quality text-image pairs, supporting both generation and evaluation tasks. In contrast to state-of-the-art benchmarks, the \data dataset offers several notable advantages:

\begin{itemize}[nosep, leftmargin=*]
    \item It represents the most comprehensive dataset to date for evaluating AI-generated images, expanding the data volume by approximately \textbf{100 times} compared to existing benchmarks.
    \item It uniquely integrates prompts from both \textbf{unimodal (text-only or image-only)} and \textbf{multi-modal (text-image combined)} inputs, effectively addressing the prevalent limitation of missing multi-modal context in prior datasets.
    \item It encompasses a \textbf{broad and diverse range of generative models}, spanning from early GAN-based architectures to the latest state-of-the-art diffusion techniques, including high-impact commercial systems such as OpenAI’s DALL·E 2 and DALL·E 3.
\end{itemize}

The extensive scale and diversity of the \data dataset provide a robust foundation for systematically analyzing the differences between natural and AI-generated images under various conditions and prompt modalities. By encompassing a wide spectrum of image styles, resolutions, semantic content, and textual descriptions, the dataset facilitates rigorous and holistic evaluations. This, in turn, supports the development and benchmarking of more precise, robust, and trustworthy AI image generation systems for real-world deployment.

\subsection{\benchmark Framework}
On top of the \data dataset, we construct the fine-grained \benchmark to systematically measure and interpret the differences between AI-generated images and realistic natural images from five key different aspects as following aspects.

\paragraph{1. Naive Image Quality Discrepancy:} We leverage traditional image quality assessment methods for natural images to measure the AI-Natural image differences. In order to perform fine-grained analysis and exploration, we estimate the AI-Natural image differences from three levels by considering the inherent properties of image-style content, ranging from the \textbf{structure-level visual features, frame-level visual features to the holistic content distribution}. 
\begin{itemize}[leftmargin=*, nosep]

    \item First, we use the full-referenced \textbf{structure-level similarity measures} like SSIM, LPIPS, and DISTS to compute the \textbf{visual similarities} between the paired AI-generated image and natural image and use these calculated values to quantify the AI-Natural image differences from the low-level visual view.

    \item Second, we use non-referenced \textbf{frame-level visual features} to qualify the perceptional quality of images and compute the frame-level visual quality difference rates between AI-generated images and natural images from the high-level visual view. The exploited frame-level visual features include PIQE \cite{ncc/NDBCM15_piqe}, IL-NIQE \cite{tip/ZhangZB15_il-niqe}, MUSIQ \cite{iccv/KeWWMY21_musiq}, DBCNN \cite{tcsv/ZhangMYDW20_dbcnn}, CLIPIQA \cite{icml/CLIPIQA}, CNNIQA~\cite{cvpr/KangYLD14_cnniqa}, TReS \cite{tmm/GuZY015_tres}, HyperIQA \cite{cvpr/SuYZZGSZ20_hyperiqa}, LIQE \cite{tip/MaLZDWZ18_liqe}, UNIQUE \cite{tip/ZhangMZY21_unique}, BRISQUE~\cite{tip/MittalMB12_brisque} , NIQE \cite{spl/MittalSB13_niqe}, NRQM \cite{tip/MaLZDWZ18_liqe}, QAlign~\cite{icml/qalign}, Inception Score (IS) \cite{nips/SalimansGZCRCC16}, etc. The Discrepancy Rate ($DR$) of these metrics provides insights into frame-level visual quality differences.
    
    \item Third, we investigate the full-reference \textbf{visual content distribution} to capture the holistic differences between AI-generated images and natural images by computing the FID, CLIP-FID, and IS of the AI-generated image set and the referenced natural image set. The results of these set-level measures can reflect the difference between AIGIs and natural ones from an integral perspective.
\end{itemize}

Based on the different levels of image quality measures, our framework can assess and measure the fine-grained AI-Natural images differences with Full-reference and Non-reference angles, covering both low-level and high-level. More details about the utilized traditional image quality assessment measures are described in Appendix~\ref{sec_app:eval_metric}.

\paragraph{2. Semantic Alignment Discrepancy:} Generative models require the referenced prompts as semantic guidance to generate images. Therefore, semantic alignment can be an important quality indicator of the AIGIs. We use the widely used CLIP model to calculate the difference rates of CLIP Scores \cite{emnlp/CLIPScore}, $DR_{CS}$ to represent alignment discrepancies between AIGIs and natural images, and provide a fine-grained analysis with respect to both unimodal and multimodal guidance.

\paragraph{3. Aesthetic Appeal Discrepancy:} Aesthetic Appeal is to estimate the visual appeal and artistic quality of images, reflecting the visual attractiveness and artistic quality of images. We utilize the classical aesthetic measures NIMA \cite{tip/TalebiM18_nima}, LAION-AES \cite{nips/laion5b}, and QAlign \cite{icml/qalign} as the quantitative metrics and compare the difference rates of aesthetic scores $DR_{aes}$ for the paired AI-generated and natural images.

\paragraph{4. Downstream Applicability Discrepancy:} This aspect is devised to investigate the practical utility of AI-generated images in downstream tasks and evaluate whether AI-generated images can have different practical utilities with realistic natural images in downstream application tasks. We mainly focus on three common downstream tasks, e.g., image recognition, object segmentation and Multimodal Visual Question Answering (VQA). For the image recognition task, we evaluate the Difference recognition rates ($DR_{re}$) between AI-generated images and natural images when using a pre-trained image recognizer like ResNet-152 \cite{cvpr/HeZRS16} model. For the object segmentation, we use the Intersection over Union (IoU) \cite{ijcv/EveringhamGWWZ10} as quantitative measures and investigate the segmentation discrepancy for AI-generated images with U\textsuperscript{2}Net \cite{pr/U2NET}. For VQA, we design three types of questions: `object' (identifying objects in the image), `alignment' (determining if the image aligns with a textual description, returning `yes' or `no'), and `similarity probability' (calculating the alignment probability between text and image). Using BLIP2 \cite{icml/blip2}, we evaluate the \textit{Difference Rate ($DR$)} between AI-generated images and reference natural images for each question type.

\paragraph{5. Human Validation:} We involve human assessments to coordinate the above evaluation aspects. We develop the human assessment interface to demonstrate AI-generated images with the referenced natural images and text descriptions and collect human ratings (on a scale from 1 to 5) alongside the following three aspects: naive image quality, semantic alignment, and Aesthetic Appeal as human assessments. Human participants are provided with example images before the evaluation to understand high and low scores without being given specific numerical values, ensuring unbiased and informed ratings. Detailed information about human assessment procedures is provided in Sec. \ref{sec:human_eval}.

Through the integration of diverse and comprehensive evaluated dimensions, our \benchmark offers a systematic assessment solution to investigate and interpret the differences that still remain between AI-generated images and natural images. 
\begin{figure*}[htp]
    \centering
    \includegraphics[width=0.9\textwidth]{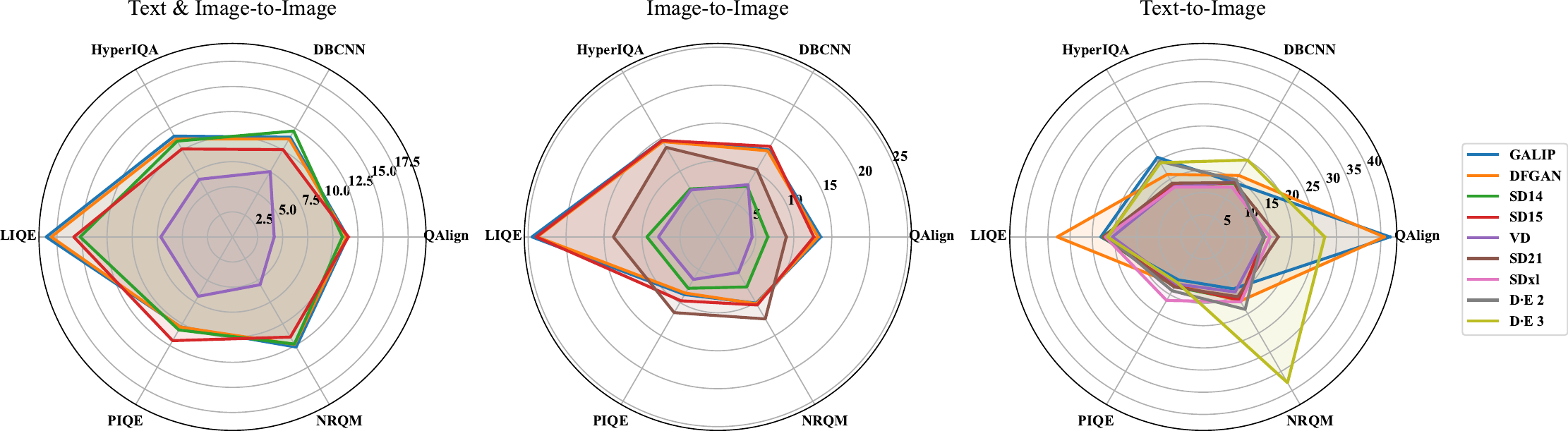}
    \caption{We evaluate nine image-generating models from 6 aspects. The numerical values in the radar chart represent the mean difference rate of each model.}
    \label{fig_radar}
\end{figure*}

\section{Evaluation}
With respect to the 5 different aspects, we conducted benchmark experiments and performed experimental evaluation alongside the following research questions.

\textbf{RQ1:} What are the fine-grained naive image quality discrepancies between AIGIs and natural images?

\begin{itemize}[nosep, leftmargin=2.3em]

    \item \textbf{RQ1.a:} What discrepancies between I2I-guided AIGIs and their natural counterparts can be interpreted via the structure-level quality measures?
    \item \textbf{RQ1.b:} What discrepancies between multimodal-guided AIGIs and their natural counterparts can be revealed via the frame-level metrics? 
    \item \textbf{RQ1.c:} How do the structural visual content distributions differ between AIGIs and natural images?   
\end{itemize}

\textbf{RQ2:} How significant are the discrepancies in the semantic alignment of the given text with natural images and AIGIs in different types of guidance?

\textbf{RQ3:} What are the discrepancies in aesthetic appeal between AIGIs and natural images?

\textbf{RQ4:} How do AIGIs differ from natural images in downstream task applicability?

\textbf{RQ5:} Are human assessment results consistent with quantitative measures? What are the discrepancies revealed from human evaluation?

\subsection{Experimental Setting}
We use the described quantitative evaluation metrics, which can be divided into Full-reference and No-reference, to present the $DR$ for involved assessment dimensions (more details of all metrics are presented in Appendix.~\ref{sec_app:eval_metric}.

\paragraph{Full-reference metrics:} For full-reference metrics, which are used for structure-level image quality, visual content distribution, and downstream object segmentation task, i.e., IoU, we directly report the calculated value to reflect the difference cause it already can present the discrepancy.

\paragraph{No-reference metrics} We calculate \textit{Difference Rate (DR)} for each quantitative metric in non-reference scenarios, by subtracting the value of each AI-generated image from its corresponding natural reference image and then averaging these difference rates, defined as follows:

\begin{equation}
    \text{\textit{DR}} = \frac{1}{n} \sum_{i=1}^{n} \frac{\lvert ((M(\mathbf{X}_i)-m_{\text{min}}) - (M(\mathbf{N}_i)-m_{\text{min}}))\rvert }{m_{\text{max}}-m_{\text{min}}},
\end{equation}
where $n$ is the number of images, $M$ represents the metric, $\mathbf{X}$ denotes the generated images, and $\mathbf{N}$ refers to the referenced natural images. Additionally, $m_{\text{min}}$ and $m_{\text{max}}$ represent the minimum and maximum values across all images, respectively, especially when the value range of the corresponding metric is undefined.

\subsection{RQ1: Naive Quality Results}
To evaluate the image quality of AI-generated images versus natural images, we conducted a comprehensive analysis using a variety of metrics that assess the discrepancies between AIGIs and natural images across structure-level similarity, frame-level quality and visual content distribution. \textbf{\textit{Our findings reveal substantial discrepancies between AIGIs and natural images across these dimensions}}.

\subsubsection{RQ1.a Structure-level}
We evaluated the visual similarity between AI-generated and natural images using structural metrics like SSIM, LPIPS, DISTS, PSNR, etc. The mean similarity values of these metrics, as shown in Table \ref{tab:similar}, indicate that \textbf{\textit{AI-generated images exhibit significantly lower similarity to natural counterparts}}. AI-generated images show a 30\% to 65\% reduction in SSIM, highlighting a major loss in structural fidelity. LPIPS and DISTS further reveal notable perceptual dissimilarities, with LPIPS scores ranging from 0.19 to 0.61. Additionally, PSNR values between 12.29 and 23.11 suggest that AI-generated images have higher noise levels, reducing their overall visual similarity.

\begin{table}[ht]
\centering
\small
\caption{Full-reference Evaluation Results (I2I).}
\label{tab:similar}
\renewcommand{\arraystretch}{1.1}
\setlength\tabcolsep{3.6pt}
    \begin{tabular}{ccccccc}
    \toprule
    Metrics & SD14   & SD15   & VD     & SD21   & SDxl   & D·E 2 \\ \midrule
    SSIM ↑  & 0.35   & 0.35   & 0.46   & 0.43   & 0.70   & 0.37     \\
    PSNR ↑  & 13.72  & 13.71  & 16.41  & 15.86  & 23.11  & 12.29    \\
    VIF ↑   & 0.02   & 0.02   & 0.05   & 0.04   & 0.18   & 0.03     \\
    VSI ↑   & 0.84   & 0.84   & 0.89   & 0.87   & 0.96   & 0.83     \\
    FSIM ↑  & 0.61   & 0.61   & 0.71   & 0.66   & 0.86   & 0.60     \\
    LPIPS ↓ & 0.60   & 0.60   & 0.39   & 0.52   & 0.19   & 0.61     \\
    DISTS ↓ & 0.27   & 0.27   & 0.17   & 0.23   & 0.10   & 0.24     \\
    MAD ↓   & 212.91 & 213.09 & 197.92 & 203.89 & 144.92 & 217.12   \\ \bottomrule
    \end{tabular}
\end{table}

\subsubsection{RQ1.b Frame-level}
We assessed the frame-level discrepancies of AI-generated and natural images using a suite of metrics, including DBCNN, HyperIQA, LIQE, PIQE, NRQM and QAlign. Figure~\ref{fig_radar} uses radar figures to visualize the mean difference rates from the six metrics, and indicates that \textbf{\textit{there are significant quality differences between AI-generated and natural images at the frame level}}. These deviations are typically around 10\%, irrespective of the guidance prompt used, with some instances showing quality differences as high as 40\%. This underscores the considerable gaps between AIGIs and natural images in terms of global image quality. Moreover, the results of NRQM reveal that \textbf{\textit{AI-generated images significantly differ from natural images in terms of naturalness}}. Specifically, AI-generated images show notable deviations from natural images in visual realism and adherence to natural scene statistics. The NRQM difference rates range from 5\% to 32\%, with images generated by D·E 3 exhibiting particularly high rates, underscoring the challenges AI-generated images encounter in replicating the inherent naturalness of real-world scenes. More detailed results with more metrics are presented in Appendix. \ref{sec_app:eval_results_quality} Figure~\ref{fig:radar_all}.

\begin{figure}[t]
    \centering

    \includegraphics[width=0.48\textwidth]{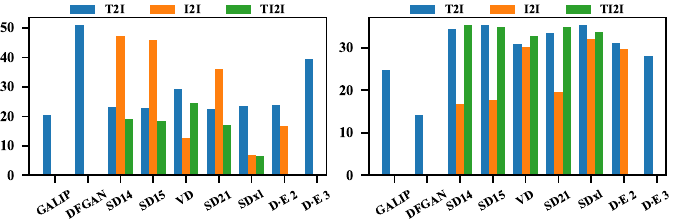}
    \caption{The FID (left, ↓) and Inception Score (right, ↑) of AI-generated images from different source models and guidance.}
    \label{fig_fid_is}
    
\end{figure}

\subsubsection{RQ1.c Visual Content Distribution}
We present the FID and Inception Score results, presented in Figure~\ref{fig_fid_is}, suggesting that AI-generated images often exhibit relatively higher FID scores (up to 50) and lower Inception Scores (low to 14). \textbf{\textit{This finding indicates that current generative models, particularly GAN-based models and earlier versions of Stable Diffusion, struggle to produce high-quality images as naturally collected images consistently.}} Notably, these models tend to generate better images when a reference image is provided (TI2I), as it has more detailed guidance.

In conclusion, \textbf{our evaluation of image quality across structure-level, frame-level, and content distribution-level reveals significant disparities between AI-generated and natural images.} These differences highlight the current limitations of AIGC technologies in matching the visual quality and realism of natural images.

\subsection{RQ2: Alignment Results}
We calculate the difference rate of CLIP Scores ($DR_{CS}$) to assess how well AI-generated images (AIGIs) align with natural images in terms of semantic consistency across three types of guidance: Text-to-Image (T2I), Image-to-Image (I2I), and Text-and-Image-to-Image (TI2I). Our results show that \textbf{\textit{AIGIs often struggle to achieve the same level of semantic alignment as natural images, particularly when guided solely by image input}}. As presented in Table~\ref{tab:aigc_alignment}, the $DR_{CS}$ ranges from 4.02\% to 36.69\%.
\textbf{This suggests that current AIGIs often face difficulties in preserving high-fidelity alignment with the provided inputs, leading to outputs that deviate from natural counterparts}. These findings highlight the ongoing challenges that AIGIs encounter in faithfully and consistently capturing the intended semantics, underscoring both their capabilities and limitations in real-world scenarios.

\begin{table}[]
    \centering
    \small
    \caption{The results of $DR_{CS}$ (\%).}
    \label{tab:aigc_alignment}  
    \renewcommand{\arraystretch}{1.1}
    \setlength\tabcolsep{2pt}
    \resizebox{1\columnwidth}{!}{
    \begin{tabular}{cccccccccc}
    \hline
         & GALIP & DFGAN & SD14  & SD15  & VD   & SD21  & SDxl  & D·E 2 & D·E 3 \\ \hline
    T2I  & 12.16 & 16.11 & 10.67 & 10.53 & 9.71 & 10.90 & 11.59 & 10.93 & 10.19 \\
    I2I  & -     & -     & 33.43 & 33.57 & 4.89 & 24.04 & 3.56  & 5.53  & -     \\
    TI2I & -     & -     & 9.90  & 9.73  & 8.22 & 9.20  & 3.57  & -     & -     \\ \hline
    \end{tabular}
    }
    \vspace{-10pt} 
\end{table}

\subsection{RQ3: Aesthetic Appeal}
\begin{figure}[t]
    \centering
    \includegraphics[width=0.48\textwidth]{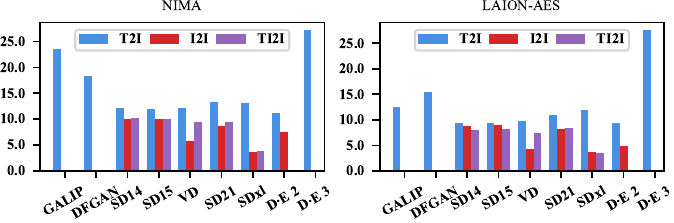}
    \caption{The $DR_{aes}$ between AIGIs and natural ones.}
    \label{fig_aesthetic}
    
\end{figure}

We assessed the aesthetic appeal of AIGIs and natural images using two widely adopted metrics: NIMA and LAION-AES. Our results indicate that \textbf{\textit{AI-generated images generally fall short of natural images in terms of aesthetic quality}}. As shown in Figure~\ref{fig_aesthetic}, the aesthetic difference rate ($DR_{aes}$) ranges from 3.51\% to 27.66\%, highlighting noticeable perceptual discrepancies between AIGIs and their natural counterparts.
These findings suggest that although current AI models are capable of generating visually appealing content, \textbf{\textit{AIGIs often lack the nuanced artistic detail and emotional resonance that are characteristic of natural images}}. This aesthetic gap underscores the limitations of existing generative systems in replicating the subtle visual cues and expressive depth found in human-created imagery.

\begin{table}[ht]
    \centering
    \small
    \caption{Image Recognition Discrepancy Rate ($DR_{re}$).}
    \label{tab:aigc_classification}    
    \renewcommand{\arraystretch}{1.1}
    \resizebox{1\columnwidth}{!}{
    \begin{tabular}{cccccccccc}
    \toprule
         & GALIP & DFGAN & SD14  & SD15  & VD    & SD21  & SDxl  & D·E 2 & D·E 3   \\ \midrule
    T2I  & 68.60 & 85.46 & 66.14 & 65.50 & 68.12 & 64.88 & 64.88 & 65.36    & 69.26 \\
    I2I  & -     & -     & 94.29 & 94.07 & 40.90 & 83.30 & 27.96 & 47.98    & -     \\
    TI2I & -     & -     & 61.62 & 59.71 & 60.05 & 55.54 & 27.46 & -        & -     \\ \bottomrule
    \end{tabular}
    }
    
\end{table}

\begin{table}[ht]
 \centering
    \small
    \caption{IoU of Object Segmentation.}
    \label{tab:aigc_segmentation}
    \renewcommand{\arraystretch}{1.1}
    \resizebox{1\columnwidth}{!}{
    \begin{tabular}{cccccccccc}
    \toprule
     & GALIP & DFGAN & SD14 & SD15 & VD & SD21 & SDxl & D·E 2 &  D·E 3 \\ \midrule
    T2I  & 0.23  & 0.24  & 0.26    & 0.25    & 0.26      & 0.25    & 0.26   & 0.26   & 0.28   \\
    I2I  &-       &-       & 0.38    & 0.38    & 0.69      & 0.50     & 0.82   & 0.42   &-        \\
    TI2I &-       &-       & 0.43    & 0.43    & 0.29      & 0.53    & 0.82   &-        &-   \\ \bottomrule     
    \end{tabular}
    }    
\end{table}

\subsection{RQ4: Applicability Results} We evaluated the applicability of AIGIs across three representative downstream tasks: image recognition, semantic segmentation, and visual question answering (VQA). Our results reveal \textbf{substantial performance discrepancies between AI-generated and natural images across all tasks}, raising concerns about the practical utility of current AIGIs in real-world applications.

\subsubsection{Image Recognition} We quantified recognition inconsistencies using the Difference Rate ($DR_{re}$, \%) of predicted labels. As shown in Table~\ref{tab:aigc_classification}, $DR_{re}$ values ranged from 27.46\% to 94.29\%, indicating that AI-generated images often fail to accurately represent the intended visual concepts, leading to incorrect or unstable recognition outcomes.

\subsubsection{Semantic Segmentation} Semantic segmentation performance was evaluated using Intersection over Union (IoU), computed based on segmentation results produced by U\textsuperscript{2}Net~\cite{pr/U2NET}. Table~\ref{tab:aigc_segmentation} reports IoU values ranging from 0.33 to 0.82, with a mean IoU of 0.40. These results suggest that generated images frequently suffer from imprecise object boundaries and incorrect spatial configurations, limiting their usability for structure-level understanding tasks.

\subsubsection{Visual Question Answering} We further assessed AI-generated images in the context of visual question answering using BLIP-2~\cite{icml/blip2}, focusing on three types of questions: Object Recognition, Semantic Alignment (Align), and Similarity Probability (Prob.) (The prompt template used here please refer to Appendix~\ref{sec_app:vaq_setting}). As summarized in Table~\ref{tab:blip}, object-based questions exhibited the highest discrepancy rates, particularly under the I2I setting, reaching up to 96.41\% (SD14) and 96.95\% (SD15). Binary alignment questions (only answer "yes" or "no") yielded lower discrepancies, with an average $DR$ of 11.55\%, while similarity-based probabilistic questions underscoring substantial discrepancies compared to natural images, with an average $DR$ of 50.41\%.

\paragraph{Summary.} Although AI-generated images perform relatively well as natural ones on simpler binary tasks, their effectiveness significantly deteriorates in more complex settings that require fine-grained visual understanding, such as object recognition, spatial reasoning, and nuanced semantic interpretation. These limitations currently hinder the broader adoption of generative images in downstream vision tasks. Our findings highlight an urgent need for further research into improving the structural fidelity and semantic coherence of generated content to enable trustworthy and task-reliable deployment.

\begin{table}[ht]
\centering
    \small
    \caption{The \textit{Difference Rate (DR,\%)} of VQA tasks.} 
    \label{tab:blip}   

    \renewcommand{\arraystretch}{1.1}
    \setlength\tabcolsep{2pt}
    \resizebox{1\columnwidth}{!}{
    \begin{tabular}{ccccccccccc}
    \toprule
                            &      & GALIP & DFGAN & SD14 & SD15 & VD & SD21 & SDxl & D·E 2  & D·E 3 \\ \midrule
    \multirow{3}{*}{\rotatebox{90}{Object}} & T2I    & 73.34  & 88.43  & 68.72   & 68.63   & 74.21     & 68.34   & 66.93  & 57.85  & 64.07  \\ 
     & I2I    & - & - & 96.41   & 96.95   & 50.43     & 91.14   & 51.65  & 46.29  & -       \\ 
     & TI2I   & - & - & 67.22   & 66.65   & 66.29     & 67.22   & 49.20  & - & -       \\ \midrule
    \multirow{3}{*}{\rotatebox{90}{Align}}  & T2I    & 10.57  & 8.55   & 14.02   & 15.79   & 12.32     & 13.02   & 13.00  & 12.99  & 21.18  \\  
     & I2I    & - & - & 9.71    & 9.53    & 7.87      & 10.07   & 8.01   & 8.93   &  -      \\ 
     & TI2I   & - & - & 13.31   & 12.70   & 10.03     & 10.99   & 8.46   & - &  -      \\ \midrule
    \multirow{3}{*}{\rotatebox{90}{Prob.}}   & T2I    & 53.13  & 62.69  & 53.07   & 54.21   & 50.57     & 54.95   & 57.09  & 37.31  & 49.22  \\ 
     & I2I    & - & - & 61.05   & 61.56   & 39.57     & 58.67   & 43.64  & 32.81  & -       \\ 
     & TI2I   & - & - & 50.96   & 50.34   & 45.93     & 49.97   & 41.48  & - &  -      \\ \bottomrule
    \end{tabular}}
\vspace{-10pt} 
\end{table}

\subsection{Human Validation}
\label{sec:human_eval}
To rigorously examine the gap between AI-generated images and human expectations, as well as to evaluate the reliability of existing metrics in assessing AI-generated images, we conducted a comprehensive human evaluation study. Participants were asked to score images across three key dimensions: \textit{Image Quality}, \textit{Alignment}, and \textit{Aesthetic Appeal}.

\begin{itemize}[nosep, leftmargin=*]

    \item \textbf{Image Quality:} Evaluates the overall visual fidelity of the image, including clarity, sharpness, and the presence of visual artifacts.
    \item \textbf{Alignment:} Assesses the semantic consistency between the generated image and its corresponding prompt, which could be textual, visual, or a combination of both.
    \item \textbf{Aesthetic Appeal:} Measures the artistic and visual attractiveness of the image, considering aspects such as composition, color balance, and overall presentation.
\end{itemize}

Participants were instructed to rate each image on a 5-point scale (1 = poor, 5 = excellent) for each of the three evaluation aspects using a radio-button interface.

\subsubsection{Interface for Human Evaluation}
The evaluation was implemented through a custom-designed interface (please refer to Appendix.~\ref{sec_app:eval_results_human}, Figure~\ref{fig:human_evaluation}). This interface enabled participants to view AI-generated images alongside their reference prompts and assign scores accordingly. The prompts used in the evaluation were drawn from the \data dataset and were provided in three modalities: (1) text-only, (2) image-only, and (3) text–image combinations. The evaluated images were generated by four state-of-the-art models: SD21, SDxl, Versatile Diffusion, and DALL·E 2.

To support score calibration and mitigate evaluation bias, we incorporated qualitative reference examples into the interface (please refer to Appendix.~\ref{sec_app:eval_results_human},  Figure~\ref{fig:human_example}). These examples illustrated typical characteristics of both strong and weak performance across the three evaluation aspects, without disclosing specific numeric scores, thereby reducing potential anchoring effects.

\subsubsection{Evaluation Process}
A total of 58 volunteers participated in the evaluation. Each participant rated a curated set of images based on the provided prompts and scoring criteria. To assess the consistency and reliability of the responses, we included duplicated image entries within the evaluation set. By comparing the scores assigned to these repeated images, we identified inconsistent raters and excluded their data from further analysis. After this filtering process, 47 participants' responses were retained as valid samples for final analysis.

\subsubsection{Participant Demographics}
The participants ranged in age from 19 to 46 years, with a gender distribution of 60\% male and 40\% female. The majority held at least an undergraduate degree, ensuring a well-educated evaluation cohort. In addition to scoring data, we collected basic demographic information and invited participants to provide open-ended feedback on their evaluation experience, particularly regarding the perceived differences between AI-generated and natural images.

\subsubsection{Results and Observations}
\begin{figure}[ht]
    \centering
    \includegraphics[width=0.48\textwidth]{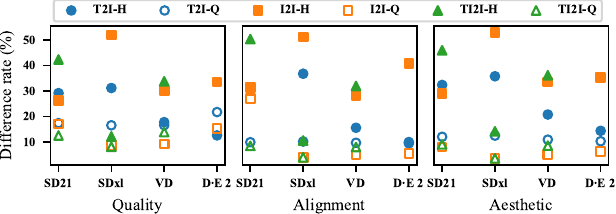}
    \caption{The difference rate provided by human \& quantitative statistics, where `H' represents human and `Q' represents quantitative.}
    \label{fig:human_quantitative}
\end{figure}

Quantitative results from the human study are summarized in Table~\ref{tab:human_study}, and visualized in Figure~\ref{fig:human_quantitative}. The findings reveal \textbf{\textit{a pronounced gap between AI-generated images and human expectations across all three evaluation dimensions}}. Specifically, AIGC images consistently received modest scores—typically between 2 and 3—indicating perceptible shortcomings in terms of visual quality, semantic fidelity, and aesthetic appeal. These results highlight the need for continued advancements in image generation technologies to better align with human perceptual standards.

Moreover, the analysis reveals that \textbf{\textit{conventional quantitative metrics often fail to reflect human perceptual judgments}} (see Figure~\ref{fig:human_quantitative}), suggesting that such metrics may be insufficient for evaluating AIGC images. This underscores the importance of developing new, human-aligned evaluation measures tailored to the unique characteristics of AI-generated visual content.

\begin{table}[h]
\centering
\small
\caption{Human scores across three aspects.}
\label{tab:human_study}
\renewcommand{\arraystretch}{1.1}
\setlength\tabcolsep{2pt}
\resizebox{1\columnwidth}{!}{
\begin{tabular}{lccccccccccc}
\toprule
\multirow{2}{*}{Aspect}         & \multicolumn{3}{c}{SD21} & \multicolumn{3}{c}{SDxl} & \multicolumn{3}{c}{VD} & \multicolumn{2}{c}{D·E 2} \\ 
          \cmidrule(lr){2-4}  \cmidrule(lr){5-7}  \cmidrule(lr){8-10} \cmidrule(lr){11-12}
          & T2I     & I2I     & TI2I    & T2I     & I2I     & TI2I   & T2I      & I2I      & TI2I    & T2I          & I2I         \\ \midrule
Quality   & 3.54    & 3.68    & 2.88    & 3.44    & 2.40     & 4.39   & 4.11     & 3.49     & 3.31    & 4.37         & 3.32        \\ \midrule
Alignment & 3.45    & 3.43    & 2.48    & 3.16    & 2.44    & 4.47   & 4.22     & 3.59     & 3.40     & 4.50          & 2.96        \\ \midrule
Aesthetic & 3.38    & 3.55    & 2.70     & 3.21    & 2.35    & 4.29   & 3.96     & 3.32     & 3.19    & 4.28         & 3.23        \\ \bottomrule
\end{tabular}
}
\end{table}

\subsection{Discrepancies Across Categories}
To systematically investigate the differences between AI-generated images (AIGI) and natural images across various content categories, we evaluate discrepancies in eight representative categories: \textit{person}, \textit{animal}, \textit{indoor}, \textit{outdoor}, \textit{vehicle}, \textit{food}, \textit{sports}, and \textit{accessory}. The evaluation is conducted along three critical dimensions: \ding{202} \textbf{Naive Quality}, assessed via structure-level similarity (SSIM), frame-level quality (CLIPIQA), and content distribution metrics (FID, CLIP-FID, and Inception Score); \ding{203} \textbf{Alignment}, measured using CLIP Score; and \ding{204} \textbf{Aesthetic Quality}, evaluated through LAION-AES.

\begin{figure}[htp]
    \centering
    \includegraphics[width=0.48\textwidth]{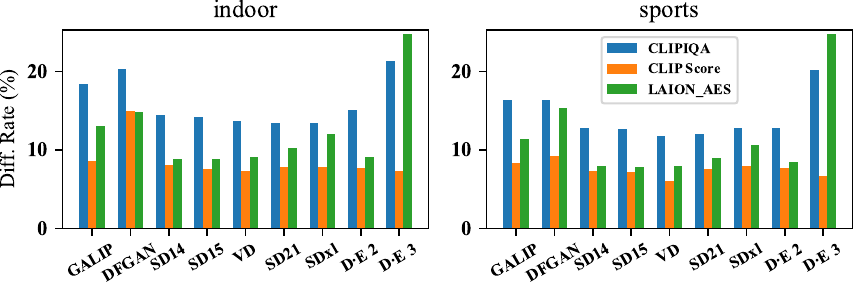}
    \caption{The difference rate on food and sport subcategories across quality, alignment, and aesthetic.}
    \label{fig:cate_nr}
\end{figure}

\begin{figure}[htp]
    \centering
    \includegraphics[width=0.48\textwidth]{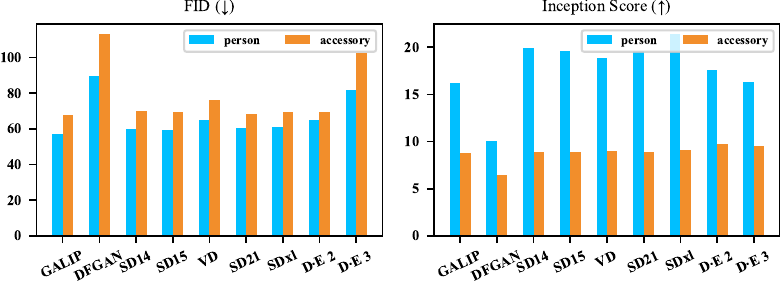}
    \caption{The FID and Inception Score (IS) of AI-generated images from different source models and categories from content distribution level.}
    \label{fig:cate_fr}
\end{figure}

Representative results are illustrated in Figures~\ref{fig:cate_nr} and \ref{fig:cate_fr}. Specifically, Figure~\ref{fig:cate_nr} presents the difference rate ($DR$) across CLIPIQA, CLIP Score, and LAION\_AES on the \textit{food} and \textit{sports} subcategories, reflecting variations in visual quality, semantic alignment, and aesthetic perception. Figure~\ref{fig:cate_fr} further provides FID and Inception Score comparisons that reveal discrepancies in content distribution for the \textit{person} and \textit{accessory} categories.

These results demonstrate that the extent of divergence between AI-generated and natural images varies significantly across categories, even when produced by the same generative model. A potential cause of this variability is the inherent imbalance in training data distributions across different semantic categories \cite{nips/laion5b,icml/AghajanyanYCHHZ23}. Additional results and visualizations are provided in Appendix~\ref{sec_app:acorss_cate}, Figures~\ref{fig:cate_nr_all} and \ref{fig:cate_fr_all}, offering a comprehensive view of these discrepancies.

Notably, Figures~\ref{fig:cate_fr_all} suggest that images generated by D·E 3 display the most pronounced deviations from natural references across multiple metrics. This could be attributed to D·E 3's inherent tendency to generate visually appealing but stylized or surreal imagery, which differs from the more realism-oriented outputs of other models (see ~\ref{sec_app:dataset}, Figure~\ref{fig:aigc_images_t2i} for visual comparisons).

\section{Conclusion}
In this paper, we present a comprehensive \benchmark based on the constructed \data dataset to evaluate the discrepancies between AI-generated and natural images. The results reveal substantial gaps across key dimensions such as image quality, semantic alignment, and downstream applicability. Despite recent progress in AIGC models, human evaluations consistently show that AI-generated images underperform in visual fidelity, semantic relevance, and aesthetic quality. Our benchmark also yields actionable guidance for improving model design, such as enhancing semantic consistency and prompt-image alignment, particularly in I2I tasks. Promising directions include reference-aware attention, perceptual alignment losses, and joint image-text encoding.

\noindent\textbf{Limitation and Future Work.}  
This study has certain limitations, including its reliance on quantitative metrics and a limited set of generative models. Future work should incorporate more diverse and balanced datasets, a broader range of contemporary models, and task-aware evaluation protocols to uncover deeper discrepancies. Beyond evaluation, \benchmark can also serve as a training signal—for instance, as a reward module in reinforcement learning or preference-based fine-tuning—to jointly optimize semantic alignment and downstream usability. We will release our dataset and code to support such use cases. Addressing these limitations and advancing model architectures will be essential for narrowing the perceptual and functional gaps between AI-generated and natural images, ultimately enabling their trustworthy deployment in real-world scenarios.

\section*{Acknowledgments}
This work is supported by Guangdong Basic and Applied Basic Research Foundation (2023A1515012848) and CCF-DiDi GAIA Collaborative Research Funds. This research is supported by A*STAR, CISCO Systems (USA) Pte.Ltd and National University of Singapore under its Cisco-NUS Accelerated Digital Economy Corporate Laboratory (Award I21001E0002). We would like to express our gratitude to all the volunteers who participated in our human evaluation experiments.

\bibliographystyle{ACM-Reference-Format}
\bibliography{ref}

\clearpage
\appendix
\section{Appendix Overview}
The appendix provides supplementary details and additional experimental results that were not included in the main paper due to space limitations. It is organized as follows:
\begin{itemize}[nosep, leftmargin=2em]
    \item \textbf{Section B:} In-depth Description of the \data Dataset.
    \item \textbf{Section C:} Comprehensive Summary of Evaluation Metrics.
    \item \textbf{Section D:} Additional Evaluation Results.
    \item \textbf{Section E:} Discussion on the Safety Mechanisms in Existing Diffusion Models.
\end{itemize}

\section{Additional Details of the \data Dataset}
\label{sec_app:dataset}

\begin{figure*}[htp]
    \centering
    \includegraphics[width=0.95\textwidth]{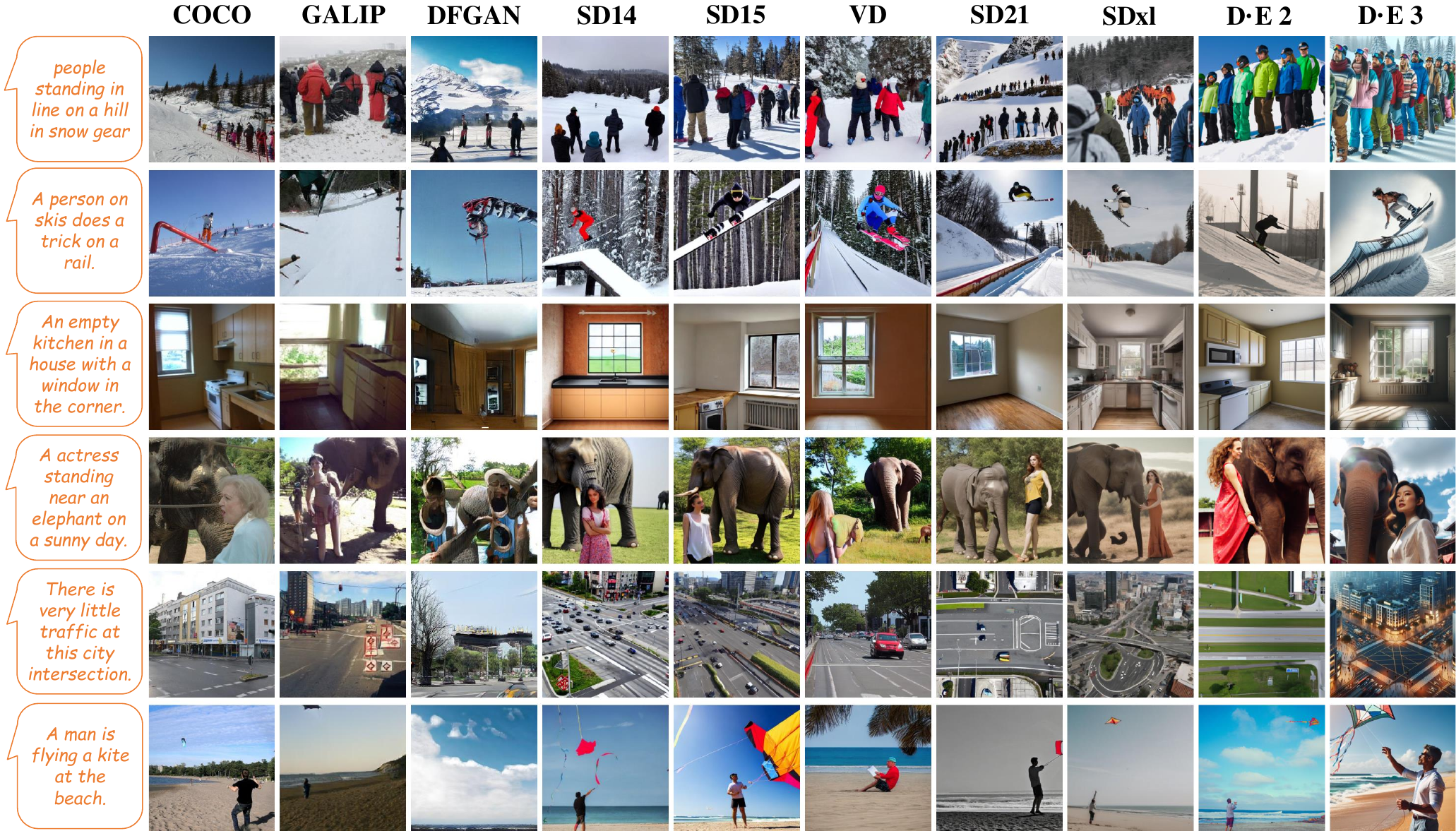}
    \caption{Examples of AI-generated content (AIGC) created using Text-to-Image (T2I) guidance.}
    \label{fig:aigc_images_t2i}
\end{figure*}

\begin{figure*}[htp]
    \centering
    \includegraphics[width=0.7\textwidth]{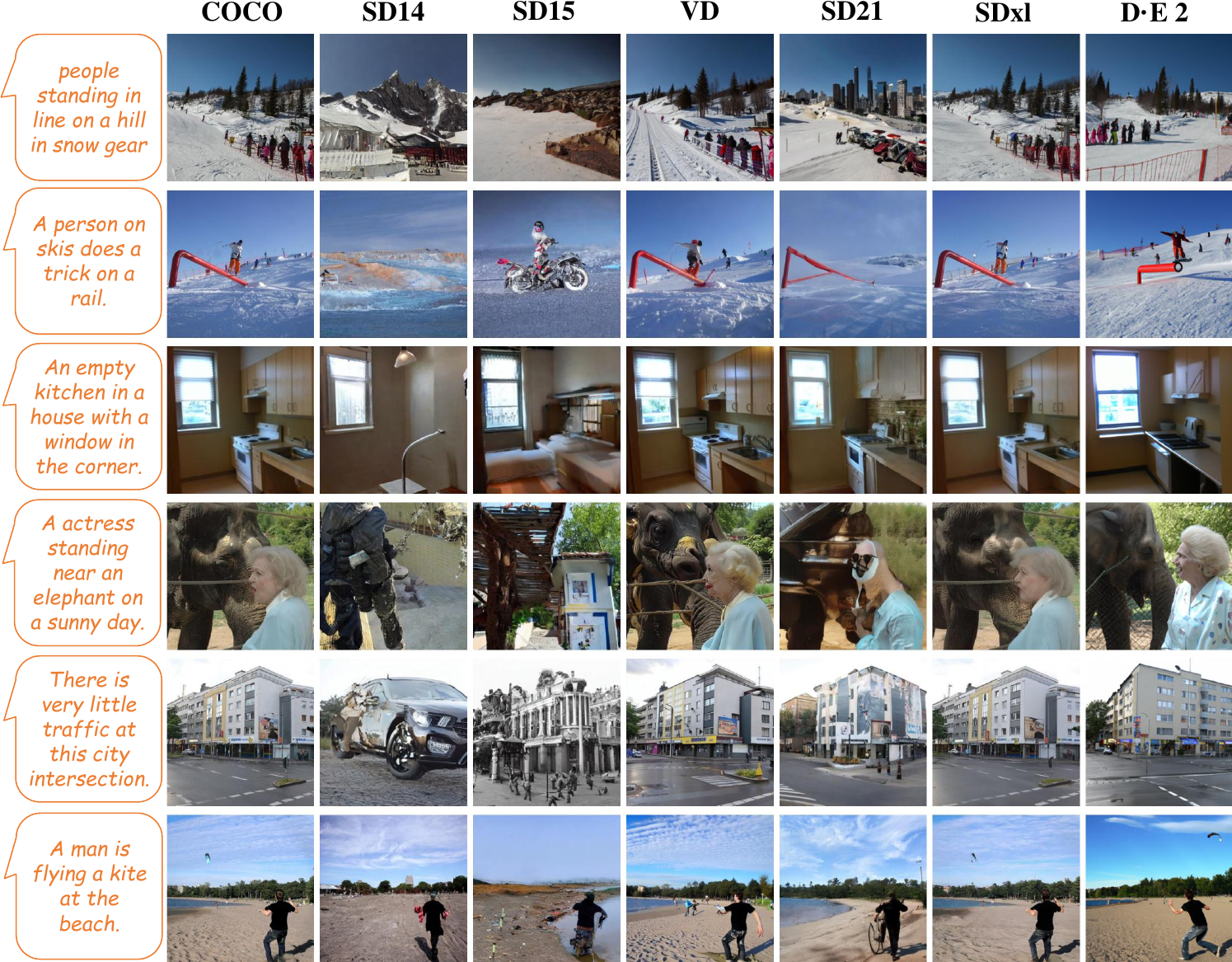}
    \caption{Examples of AI-generated content (AIGC) created using Image-to-Image (I2I) guidance.}
    \label{fig:aigc_images_i2i}
\end{figure*}

\begin{figure*}[htp]
    \centering
    \includegraphics[width=0.6\textwidth]{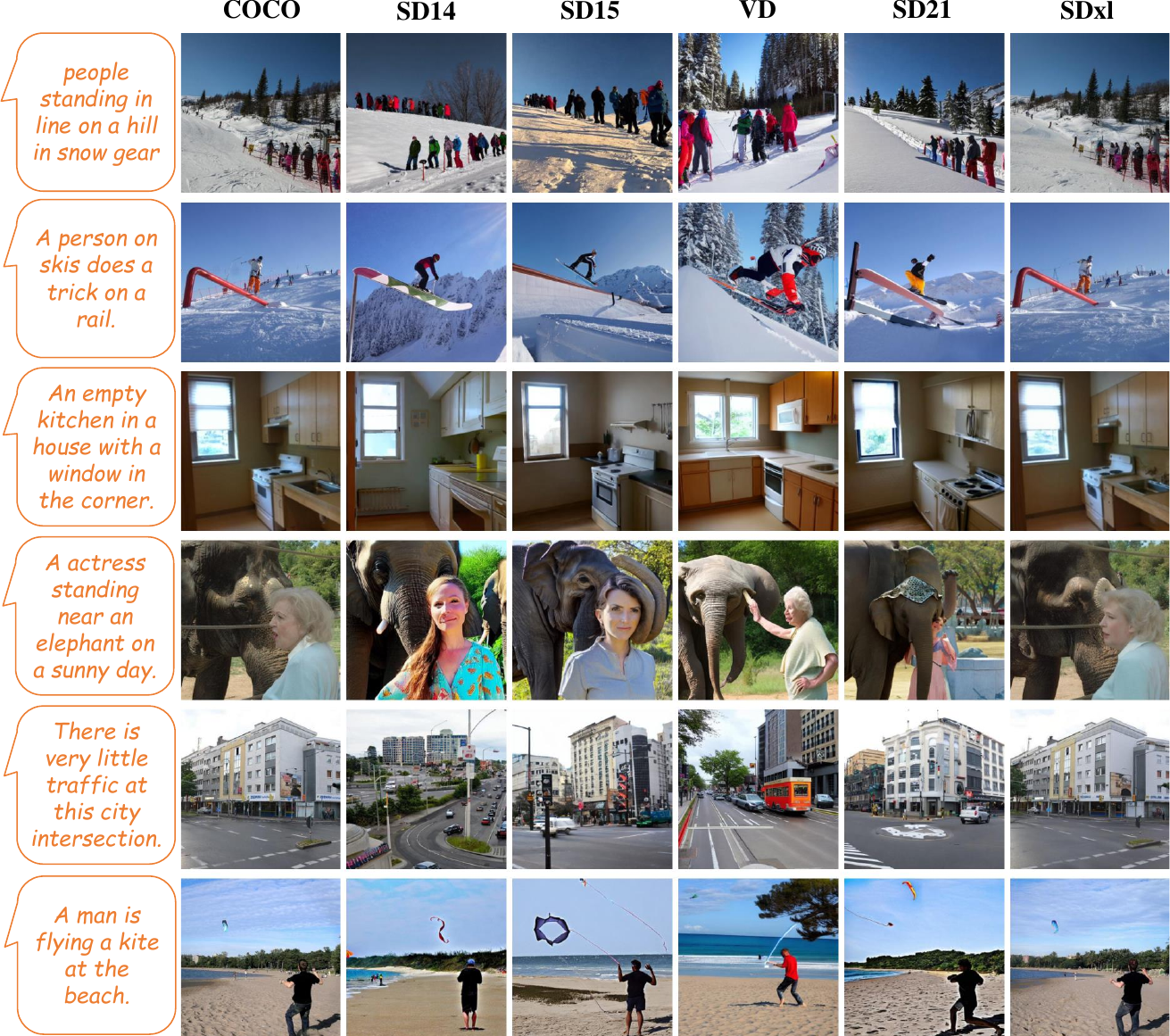}
    \caption{Examples of AI-generated content (AIGC) created using Text-and-Image-to-Image (TI2I) guidance.}
    \label{fig:aigc_images_ti2i}
\end{figure*}

The \data dataset is a carefully curated collection of both natural and AI-generated images, designed to support comprehensive evaluations of image generation models. The natural images are sourced from the COCO validation set. AI-generated images were produced using nine different models. These images were generated under three distinct guidance modes: Text-to-Image (T2I), Image-to-Image (I2I), and Text \textit{vs.} Image-to-Image (TI2I). Representative samples from the \data dataset are shown in Figures~\ref{fig:aigc_images_t2i}, ~\ref{fig:aigc_images_i2i} and ~\ref{fig:aigc_images_ti2i}. 

The involved generative models are as follows:
\begin{itemize}
    \item \textbf{GALIP (Generative Adversarial Latent Image Processing):} \cite{cvpr/TaoB0X23} is a generative model that produces high-quality images by leveraging adversarial techniques. It utilizes latent image processing to enhance both the quality and fidelity of generated content. GALIP’s architecture involves a generator and discriminator, where the generator creates images from latent vectors, and the discriminator evaluates their authenticity. Through iterative training, GALIP achieves highly detailed and realistic image synthesis.

    \item  \textbf{DFGAN (Deep Fusion Generative Adversarial Network):} \cite{cvpr/Tao00JBX22} focuses on generating images from text descriptions using deep fusion techniques. The model integrates textual information into the image generation process through multiple layers of fusion, ensuring alignment with input descriptions. DFGAN operates in two stages: the first generates a coarse image based on the text, and the second refines the image, enhancing detail and coherence.

    \item \textbf{Stable Diffusion v1.4 (SD14):} \cite{cvpr/RombachBLEO22} is part of the Stable Diffusion family, designed for image synthesis through iterative diffusion processes. Starting with pure noise, the model refines it into coherent images through a series of steps, guided by a learned model that predicts noise distribution. SD14 is recognized for producing high-resolution images with fine textures and complex structures.

    \item \textbf{Stable Diffusion v1.5 (SD15):} \cite{cvpr/RombachBLEO22} builds on SD14, introducing enhancements in architecture and training techniques to improve image quality, consistency, and the handling of complex scenes. While continuing to use the diffusion process, SD15 incorporates better optimization strategies and larger datasets, yielding superior results.

    \item \textbf{Versatile Diffusion:} \cite{iccv/XuWZWS23} is a diffusion-based model designed for a variety of image generation tasks, including style transfer, image inpainting, and super-resolution. Its versatility lies in its ability to adapt the diffusion process to the specific requirements of each task, making it suitable for diverse applications.

    \item \textbf{Stable Diffusion v2.1 (SD21):} \cite{cvpr/RombachBLEO22} is an advanced iteration of the Stable Diffusion series, offering significant improvements in image fidelity and generation speed. With optimizations in the diffusion process, better noise handling, and enhanced training algorithms, SD21 produces more realistic and high-quality images, making it highly effective for a range of creative and practical uses.

    \item \textbf{Stable Diffusion XL (SDxl):} \cite{cvpr/RombachBLEO22} represents the most advanced model in the Stable Diffusion series, capable of generating extremely high-resolution images with intricate details. SDxl uses an extended number of diffusion steps and a larger network architecture to manage increased complexity, making it ideal for applications requiring ultra-high resolution and precision, such as detailed artworks and large-scale prints.

    \item \textbf{DALL·E 2 (D·E 2):} \cite{corr/abs-2204-06125} is a generative model by OpenAI designed to create images from textual descriptions. It produces high-quality, diverse, and coherent outputs by combining CLIP (Contrastive Language-Image Pre-training) with a transformer-based generative architecture. DALL·E 2 excels at translating complex and abstract textual prompts into visually captivating content, establishing itself as a benchmark model in text-to-image generation for its creativity and realism.

    \item \textbf{DALL·E 3 (D·E 3):} \cite{dalle3} is the latest evolution in OpenAI's text-to-image generative models, offering significant advancements over DALL·E 2. It excels in interpreting nuanced and detailed prompts, producing outputs that are more coherent, visually appealing, and aligned with the given descriptions. DALL·E 3 integrates seamlessly with GPT-4, leveraging its advanced language understanding to ensure a deeper alignment between textual input and generated images. This integration enhances both the quality and controllability of image generation, making DALL·E 3 a powerful tool for a wide range of creative and professional applications.
\end{itemize}

These generative models represent significant advancements in AI-driven image synthesis, each with unique strengths tailored to specific applications. From text-to-image generation and high-resolution synthesis to versatile image processing, these models push the boundaries of AI-generated content.

\section{Evaluation Metrics}
\label{sec_app:eval_metric}

\subsection{Overview}
In this study, we selected a comprehensive set of metrics to evaluate the image quality, alignment, and aesthetics of both natural and AI-generated images. These metrics, summarized in Table \ref{tab:metrics}, are carefully chosen to highlight the differences across multiple dimensions, providing a robust framework for assessing the quality and realism of AIGC in comparison to natural images.

To evaluate image quality, we divided the assessment into three sub-aspects: structure-level similarity, frame-level visual metrics, and visual content distribution.
\begin{itemize}
    \item \textbf{Structure-level:} Metrics such as SSIM, PSNR, LPIPS, DISTS, VIF, VSI, FSIM, and MAD assess structural, perceptual, and textural fidelity. These metrics measure how closely the generated images match natural ones in terms of structural integrity and perceptual similarity.

    \item \textbf{Frame-level:} Metrics including PIQE, IL-NIQE, MUSIQ, DBCNN, CNNIQA, CLIPIQA, BRISQUE, TReS, HyperIQA, LIQE, UNIQUE and QAlign provide no-reference assessments of image quality. They evaluate attributes such as distortion levels, natural scene statistics, and learned features to predict the overall quality of the images. Additionally, naturalness metrics like NIQE and NRQM help gauge how closely AI-generated images resemble natural scenes based on statistical properties.

    \item \textbf{Visual Content Distribution:} Metrics such as FID, CLIP-FID and Inception Score examine overall content distribution differences between AI-generated and natural images across the entire dataset. These metrics capture broad differences in content quality.
\end{itemize}

For alignment, we utilize the CLIP Score \cite{emnlp/CLIPScore} to measure the correspondence between text descriptions and the generated images. This metric evaluates how well the generated content aligns with the given prompts by embedding both text and image into a shared space and calculating their similarity.

In terms of aesthetic evaluation, we employ NIMA, LAION-AES and QAlign metrics to assess the visual appeal and artistic quality of the images. These metrics predict human perception of image aesthetics based on deep learning models and various visual features.

By integrating these diverse metrics, our framework offers a detailed and multi-faceted evaluation of AI-generated images, capturing the nuances and differences between natural and synthetic content. The comprehensive nature of these metrics ensures a robust and reliable assessment, guiding improvements in AIGC technologies.

The chosen metrics, as detailed in Table \ref{tab:metrics}, collectively provide a comprehensive toolkit for evaluating image quality, alignment, and aesthetics, each focusing on different aspects of visual perception and fidelity.

\subsection{Naive Image Quality}
We evaluated the naive image quality of AI-generated images versus natural images using a range of metrics. These metrics assess both structure-level and frame-level qualities, as well as visual content distribution, to capture the holistic differences between AI-generated and natural images.

\paragraph{Structure-Level Quality Metrics:}
\begin{itemize}
    \item \textbf{SSIM} \cite{tip/WangBSS04} (Structural Similarity Index) compares the structural information between a reference and a test image, considering luminance, contrast, and structure for a comprehensive measure of similarity.
    
    \item \textbf{PSNR} (Peak Signal-to-Noise Ratio) measures the ratio between the maximum possible power of an image and the power of corrupting noise, with higher values indicating better image quality.
    
    \item \textbf{VIF} \cite{tip/SheikhB06_vif} (Visual Information Fidelity) quantifies the amount of visual information preserved in the test image relative to the reference, based on natural scene statistics and the human visual system.
    
    \item \textbf{VSI} \cite{tip/ZhangSL14_vsi} (Visual Saliency-based Index) evaluates perceptual similarity by focusing on how salient (prominent) features in the images compare.
    
    \item \textbf{FSIM} \cite{tip/ZhangZMZ11_fsim} (Feature Similarity Index) measures similarity using phase congruency and gradient magnitude, emphasizing feature similarity, which is crucial for human perception.
    
    \item \textbf{LPIPS} \cite{cvpr/LPIPS} (Learned Perceptual Image Patch Similarity) uses deep network features to assess perceptual similarity, capturing human visual perception more effectively by comparing feature activations in a deep neural network.
    
    \item \textbf{DISTS} \cite{pami/DingMWS22_dists} (Deep Image Structure and Texture Similarity) combines structural and textural similarity metrics using deep learning features to evaluate overall image similarity.
    
    \item \textbf{MAD} \cite{jei/LarsonC10_mad} (Mean Absolute Deviation) provides a straightforward assessment of structure-level differences, with lower values indicating greater similarity between images.
\end{itemize}

\paragraph{Frame-Level Quality Metrics:}
\begin{itemize}
    \item \textbf{PIQE} \cite{ncc/NDBCM15_piqe} (Perception-based Image Quality Evaluator) provides a no-reference quality assessment by analyzing image blocks and estimating distortion levels.
    
    \item \textbf{IL-NIQE} \cite{tip/ZhangZB15_il-niqe} (Integrated Local Natural Image Quality Evaluator) evaluates image quality based on natural scene statistics and local features, offering a no-reference assessment.
    
    \item \textbf{MUSIQ} \cite{iccv/KeWWMY21_musiq} (Multi-Scale Image Quality) assesses image quality at multiple scales, considering both local and global features for a comprehensive no-reference evaluation.
    
    \item \textbf{DBCNN} \cite{tcsv/ZhangMYDW20_dbcnn} (Deep Bilinear Convolutional Neural Network) predicts image quality using deep learning, based on bilinear pooling of convolutional features.
    
    \item \textbf{LIQE} \cite{tip/MaLZDWZ18_liqe} (Learning-based Image Quality Evaluator) combines traditional quality metrics with learned features for robust no-reference quality assessment.
    
    \item \textbf{CNNIQA} \cite{cvpr/KangYLD14_cnniqa} (Convolutional Neural Network Image Quality Assessment) leverages deep convolutional networks for no-reference image quality assessment using learned features.
    
    \item \textbf{CLIPIQA} \cite{icml/CLIPIQA} (CLIP Image Quality Assessment) uses CLIP model embeddings to assess image quality by comparing the alignment between image and text descriptions.
    
    \item \textbf{TReS} \cite{icml/CLIPIQA} (Textural and Edge-based Similarity) evaluates image quality by focusing on textural and edge-based features, providing a no-reference assessment.
    
    \item \textbf{HyperIQA} \cite{cvpr/SuYZZGSZ20_hyperiqa} (Hyper Network-based Image Quality Assessment) uses a hypernetwork to predict image quality, leveraging deep learning for no-reference quality assessment.
    
    \item \textbf{UNIQUE} \cite{tip/ZhangMZY21_unique} (Universal Quality Index with Deep Features) evaluates unique aspects of image quality using deep learning features, offering a no-reference assessment.
    
    \item \textbf{BRISQUE} \cite{tip/MittalMB12_brisque} (Blind/Referenceless Image Spatial Quality Evaluator) assesses image quality based on natural scene statistics, providing a no-reference measure.
    
    \item \textbf{NIQE} \cite{spl/MittalSB13_niqe} (Natural Image Quality Evaluator) evaluates how natural an image appears by comparing its features with those derived from a dataset of natural images, without needing a reference image.
    
    \item \textbf{NRQM} \cite{tip/MaLZDWZ18_liqe} (Naturalness Image Quality Metric) assesses how closely an image aligns with the statistical properties of natural scenes based on natural scene statistics and perceptual models.

    \item \textbf{QAlign} \cite{icml/qalign} evaluates image quality by assessing its alignment with semantic and perceptual features indicative of high visual fidelity.  

\end{itemize}

\paragraph{Visual Content Distribution Metrics:}
\begin{itemize}
    \item \textbf{FID} \cite{nips/FID} (Fréchet Inception Distance) measures the similarity between two sets of images by comparing the means and covariances of their feature representations, providing a comprehensive assessment of both quality and diversity.

    \item \textbf{CLIP-FID} \cite{iclr/clipfid} (CLIP Fréchet Inception Distance) is an extension of FID that measures the similarity between two sets of images by utilizing CLIP's multi-modal feature space. Instead of relying solely on Inception-based features, CLIP-FID leverages the semantic-rich embeddings from CLIP, which are aligned across image and text modalities. This provides a more nuanced assessment of image quality and semantic alignment, particularly for generative models designed to capture complex textual descriptions.
    
    \item \textbf{Inception Score} \cite{nips/SalimansGZCRCC16} evaluates the quality of generated images by considering the confidence of the classifier and the diversity of the generated samples, with higher scores indicating better quality and diversity.
\end{itemize}

\subsection{Alignment}

To assess alignment, we employ the \textbf{CLIP Score} (Contrastive Language-Image Pre-training Score, CS), which measures the correspondence between text descriptions and generated images. This metric evaluates how well the visual content aligns with the textual prompts by embedding both into a shared space and calculating their cosine similarity.

\subsection{Aesthetic}

\begin{itemize}
    \item \textbf{NIMA} \cite{tip/TalebiM18_nima} (Neural Image Assessment) is a deep learning-based model that predicts the aesthetic quality of images. It generates a score that reflects human perception of visual appeal.

    \item \textbf{LAION-AES} \cite{nips/laion5b} (LAION Aesthetic Score) evaluates images' aesthetic quality using various visual features, providing a numerical value to indicate their aesthetic appeal.

    \item \textbf{QAlign} \cite{icml/qalign} assesses image aesthetics by analyzing its adherence to human visual preferences and artistic principles, leveraging semantic coherence and appeal.
\end{itemize}

Together, these metrics offer a robust toolkit for evaluating images, focusing on alignment with prompts and aesthetic quality, thereby covering key aspects of visual perception and fidelity.

\begin{table}[ht]
\small
\centering
\caption{The detail of evaluation metrics.}
\label{tab:metrics}
\renewcommand{\arraystretch}{1.2}
\begin{tabular}{cccc}
\toprule
Aspects                               & Sub aspects                                 & Metrics & Reference \\ \midrule
\multirow{25}{*}{Naive Quality} & \multirow{8}{*}{Structure-level}                 & SSIM            & TRUE      \\
                                      &                                              & PSNR            & TRUE      \\
                                      &                                              & LPIPS           & TRUE      \\
                                      &                                              & DISTS           & TRUE      \\
                                      &                                              & VIF             & TRUE      \\
                                      &                                              & VSI             & TRUE      \\
                                      &                                              & FSIM            & TRUE      \\
                                      &                                              & MAD             & TRUE      \\ \cmidrule{2-4} 
                                      & \multirow{14}{*}{Frame-level}                & PIQE            & FALSE     \\
                                      &                                              & IL-NIQE         & FALSE     \\
                                      &                                              & MUSIQ           & FALSE     \\
                                      &                                              & DBCNN           & FALSE     \\
                                      &                                              & CNNIQA          & FALSE     \\
                                      &                                              & CLIPIQA         & FALSE     \\
                                      &                                              & BRISQUE         & FALSE     \\
                                      &                                              & TReS            & FALSE     \\
                                      &                                              & HyperIQA        & FALSE     \\
                                      &                                              & LIQE            & FALSE     \\
                                      &                                              & UNIQUE          & FALSE     \\
                                      &                                              & NIQE            & FALSE     \\
                                      &                                              & NRQM            & FALSE     \\ 
                                      
                                      &                                              & QAlign (quality)       & FALSE     \\\cmidrule{2-4} 
                                      & \multirow{3}{*}{Content Distribution} & FID             & TRUE      \\
                                      
                                      &                                              & CLIP-FID & TRUE      \\
                                      
                                      
                                      &                                              & Inception Score & TRUE      \\ \midrule
Alignment                             &                                              & CLIP Score     & FALSE     \\ \midrule
\multirow{3}{*}{Aesthtic}             & \multirow{3}{*}{}                            & NIMA            & FALSE     \\
                                      &                                              & LAION-AES       & FALSE     \\
                                      &                                              & QAlign (aesthetic)       & FALSE     \\ \bottomrule
\end{tabular}

\end{table}

\begin{figure*}[ht]
    \centering
    \includegraphics[width=0.9\textwidth]{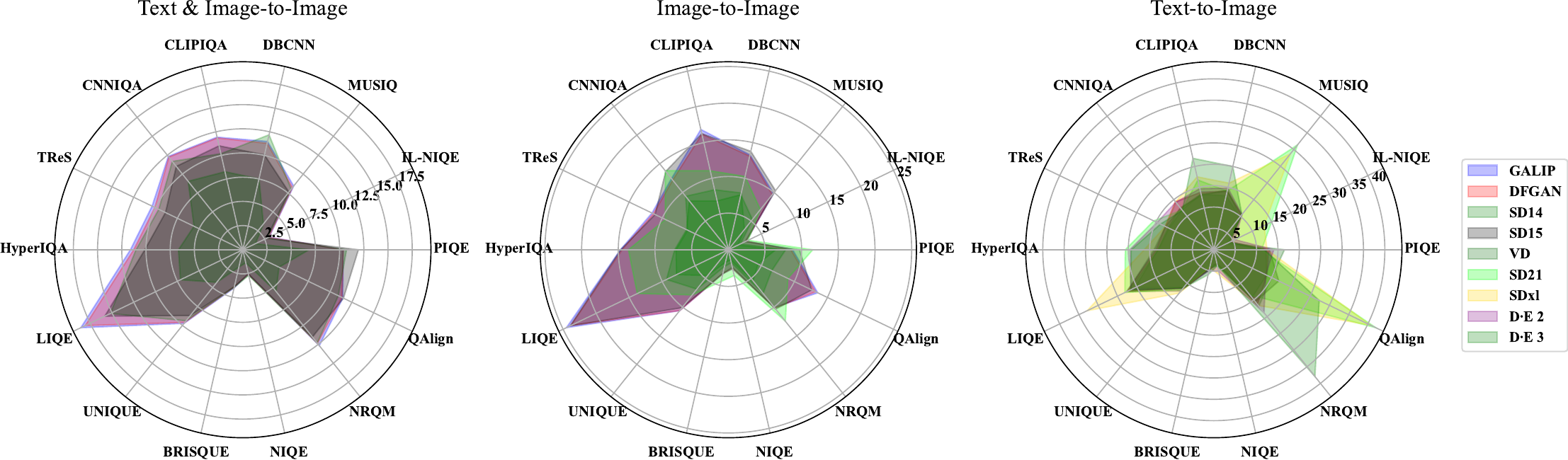}
    \caption{We evaluate nine image-generating models from 14 aspects. The numerical values in the radar chart represent the mean difference rate of each model and each metric.}
    \label{fig:radar_all}
\end{figure*}

\section{Evaluation}
\label{sec_app:eval_results}

\subsection{Naive Quality}
\label{sec_app:eval_results_quality}
We present the additional results of the Naive Image Quality assessment in Figure~\ref{fig:radar_all}.

\subsubsection{Aesthetic Appeal}
\label{sec_app:aesthic}
We present the additional results of QAlign~\cite{icml/qalign} regarding to the Aesthetic Appeal assessment in Table~\ref{tab:qalign_aes}.

\begin{table}[h]
 \centering
    \small
    \caption{Aesthetic Appeal of QAlign.}
    \label{tab:qalign_aes}
    \renewcommand{\arraystretch}{1.1}
    \resizebox{1\columnwidth}{!}{
    \begin{tabular}{cccccccccc}
\toprule
     Types & DFGAN & GALIP & SD14 & SD15 & VD & SD21 & SDxl & D·E 2 & D·E 3 \\ \midrule
T2I  & 45.33     & 39.35     & 32.44 & 31.84 & 35.31 & 40.91 & 43.73 & 26.53 & 99.34 \\
I2I  & -         & -         & 23.47 & 23.09 & 11.31 & 20.32 & 8.17  & 13.88 & -     \\
TI2I & -         & -         & 25.8  & 25.98 & 23.76 & 26.62 & 7.46  & -     & -     \\ \bottomrule
\end{tabular}
    }    
\end{table}

\subsection{Applicability Settings}
\label{sec_app:vaq_setting}
To facilitate the applicability evaluation in downstream tasks, we construct three standardized Visual Answering Question (VAQ) templates using BLIP-2, each targeting a specific aspect of visual-semantic understanding: object recognition, semantic alignment, and similarity probability. As shown in Table~\ref{tab:VAQ_questions}, each prompt is designed to elicit structured and interpretable responses. The object recognition prompt asks the model to identify the main object in the image and return a class label from the COCO taxonomy. The alignment prompt evaluates whether the image content semantically matches a given caption, requiring a binary “yes” or “no” answer. The similarity probability prompt estimates the degree of alignment between image and caption by requesting a numerical probability. These prompts are uniformly applied to both AI-generated and reference images, offering a consistent protocol for probing semantic fidelity across different generation conditions.

\begin{table}[]
\small
\centering
\caption{The prompt template used for VAQ tasks.}
\label{tab:VAQ_questions}
\begin{tabular}{l p{5cm}} 
\toprule
Task                   & Prompts                                                                                                                  \\
\midrule
Object                 & \textbf{Question}: what is the object in the image? Only return the class name in \{COCO class list\}. \textbf{Answer}:                  \\
Alignment              & \textbf{Question}: the image content aligns very well with \{caption\}? only return 'yes' or 'no'. \textbf{Answer}:                      \\
Similarity probability & \textbf{Question}: what's the probability of the image content well align with \{caption\}? only return a float number. \textbf{Answer}: \\
\bottomrule
\end{tabular}
\end{table}

\subsection{ Human Perceptual Evaluation}
\label{sec_app:eval_results_human}
\begin{figure*}[ht]
    \centering
    \includegraphics[width=\textwidth]{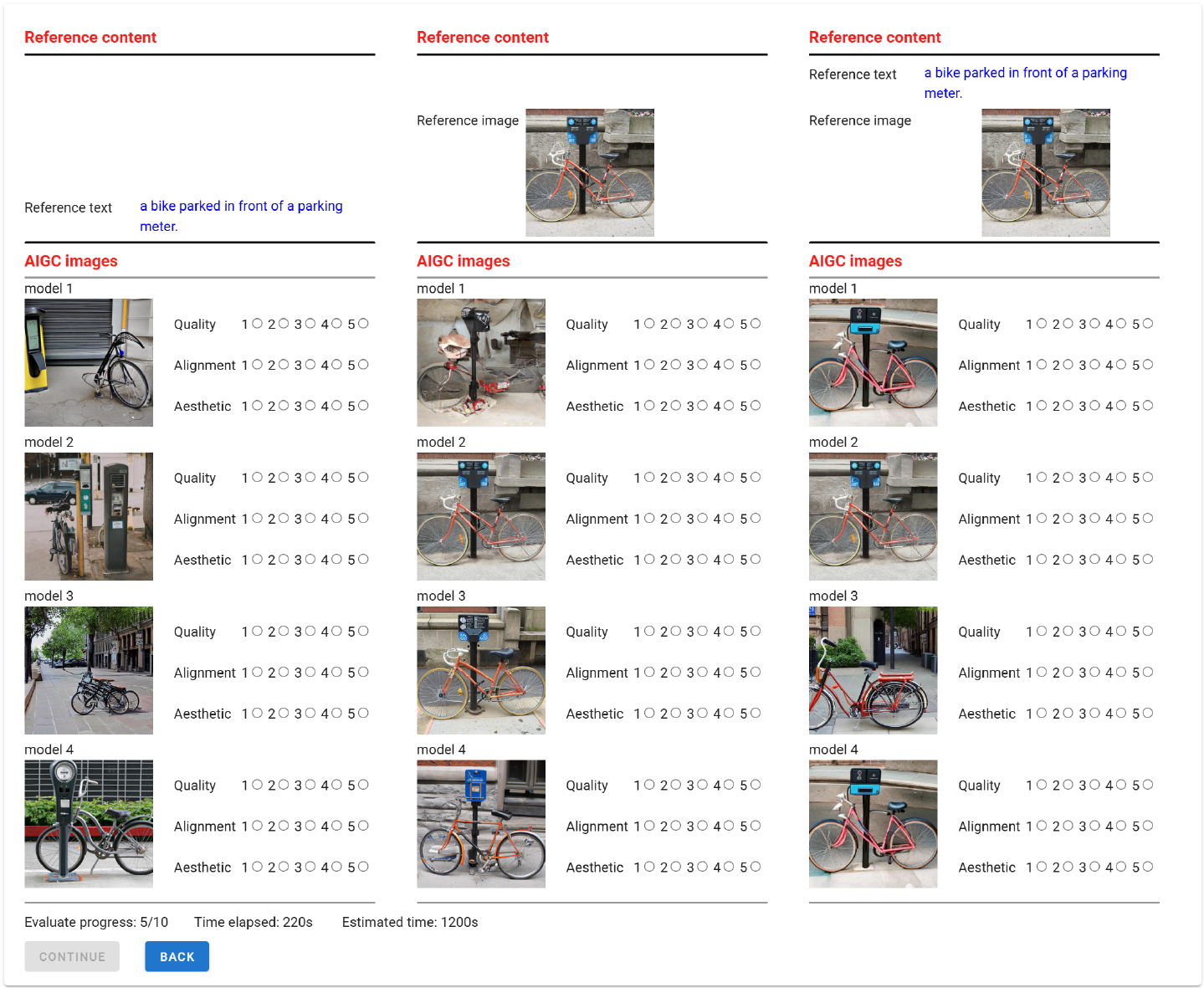}
    \caption{Interface for Human Evaluation. Reference content: the prompt guidance used to generate images, including three types: 1) text, 2) image, and 3) text and image. AIGC images: AI-generated images that come from four different models. Participants can rely on the reference content to evaluate each image on three aspects, i.e., Quality, Alignment, and Aesthetic.}
    \label{fig:human_evaluation}
\end{figure*}

\begin{figure*}[ht]
    \centering
    \includegraphics[width=\textwidth]{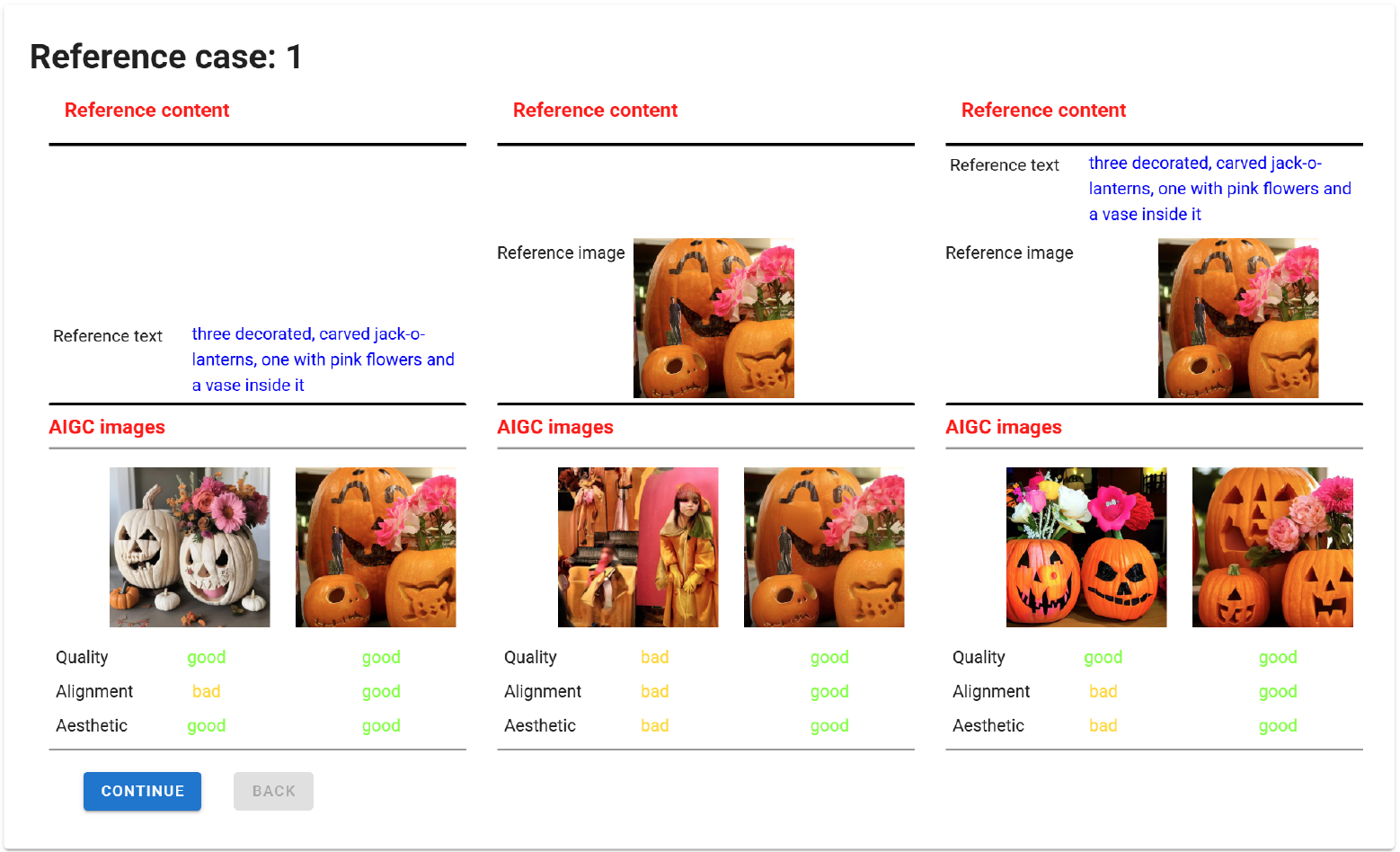}
    \caption{Before the evaluation process, the participant will be given some examples to learn which image should be good or bad in a specific aspect.}
    \label{fig:human_example}
\end{figure*}

To complement the description in Sec.\ref{sec:human_eval}, we present two figures that detail the setup of our human evaluation protocol. Figure\ref{fig:human_evaluation} illustrates the custom-designed interface through which participants evaluated AI-generated images. The interface supports side-by-side comparisons and collects ratings across three key dimensions: Quality, Alignment, and Aesthetics. For each dimension, participants assigned a score ranging from 1 to 5 using a radio-button selection mechanism, where higher scores indicated better performance.
Figure~\ref{fig:human_example} shows the set of reference examples embedded within the interface. These examples were designed to convey representative cases of both high and low performance for each evaluation dimension. They served as non-numeric visual guidelines to help participants develop a consistent evaluation criterion, while avoiding explicit scoring anchors that might introduce bias.

\begin{figure}[ht]
    \centering
    \includegraphics[width=0.48\textwidth]{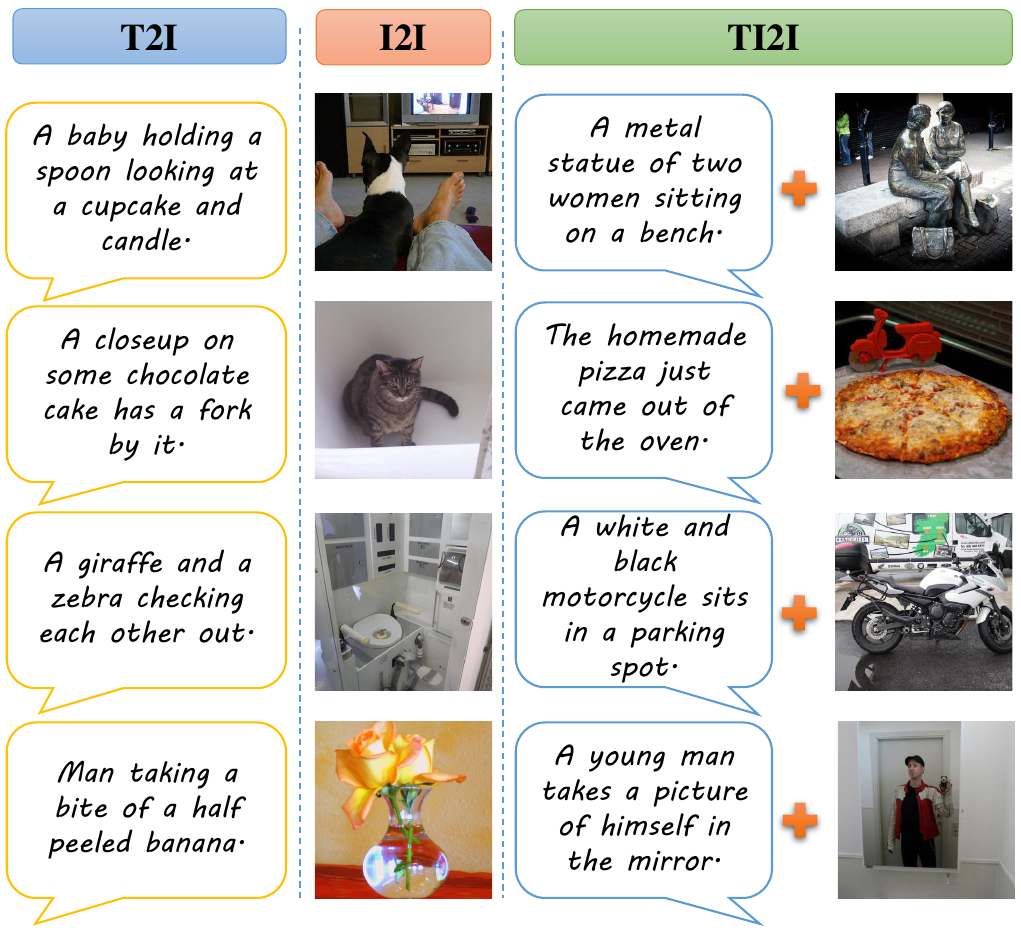}
    \caption{Examples for unsafe references.}
    \label{fig:unsafe}
\end{figure}

\begin{figure*}[!t]
    \centering
    \includegraphics[width=0.98\textwidth]{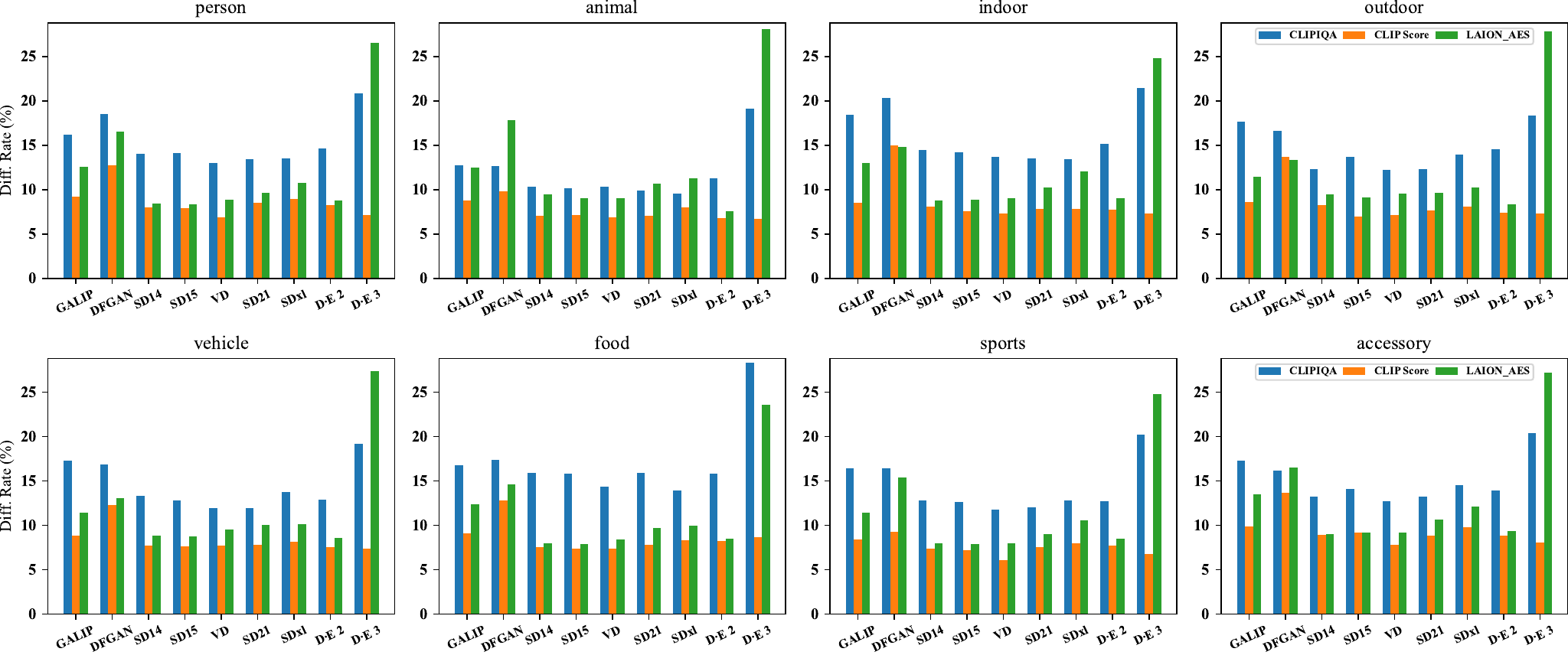}
    \caption{The differential rate of data from different categories models across naive image quality, alignment, and aesthetic.}
    \label{fig:cate_nr_all}
\end{figure*}

\begin{figure*}[!t]
    \centering
    \includegraphics[width=0.82\textwidth]{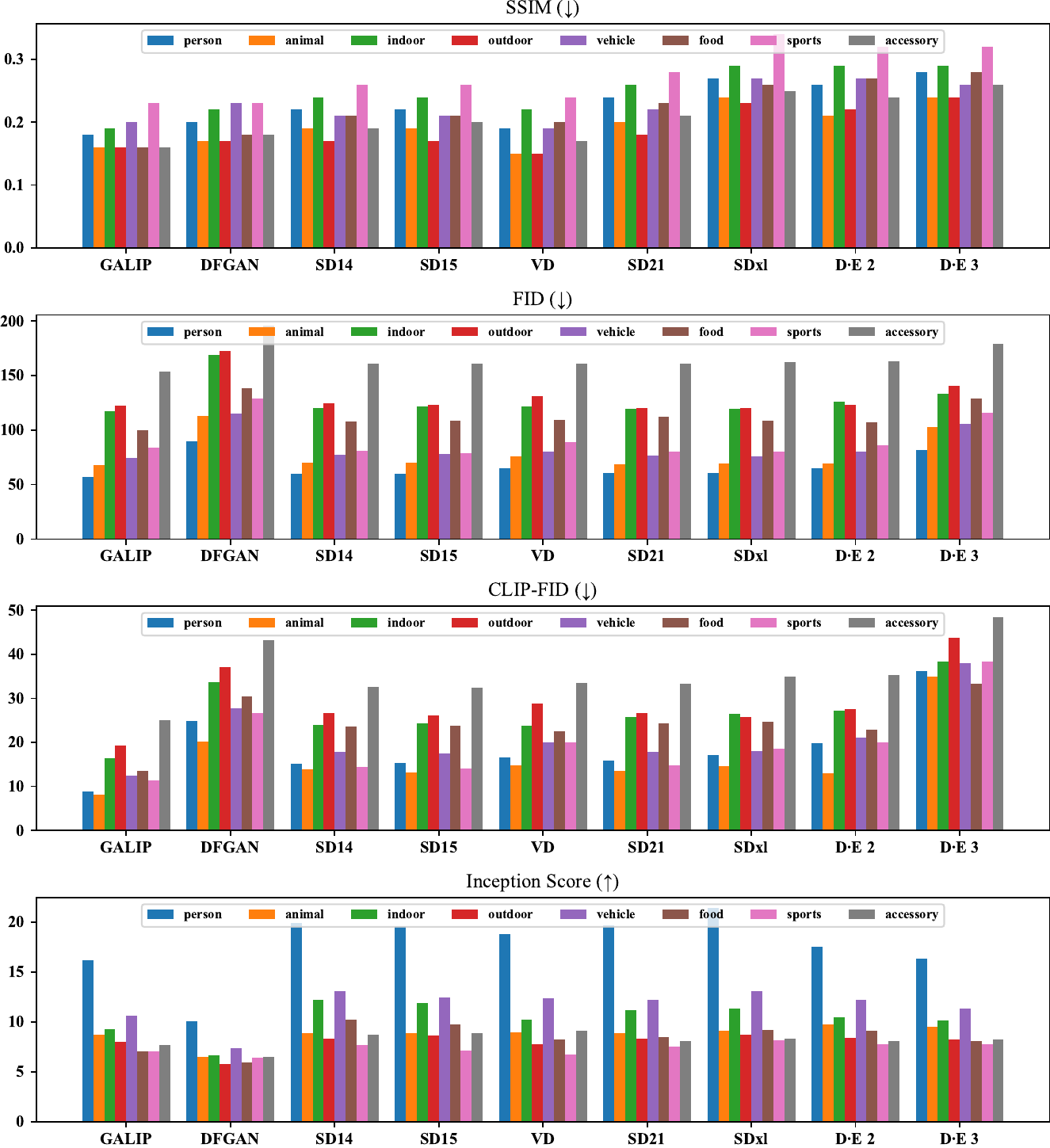}
    \caption{The FID, CLIP-FID, and Inception Score (IS) of AI-generated images from different source models and categories from the content distribution level.}
    \label{fig:cate_fr_all}
\end{figure*}

\subsection{Discrepancies Across Categories}
\label{sec_app:acorss_cate}
This section provides a detailed evaluation of AI-generated images across different categories, focusing on naive image quality, alignment, aesthetics, structure-level similarity, and content distribution. These analyses aim to uncover discrepancies between AI-generated and natural images and identify category-specific challenges.

From Figure~\ref{fig:cate_nr_all}, we observe notable variability in the differential rates across categories such as `indoor' and `sports' among different generative models. Images in the `indoor' category exhibit significantly higher discrepancies in alignment (CLIP Score) and aesthetics (LAION-AES) compared to `sports', indicating that AI-generated sports images are more aligned with natural images. In contrast, generating natural-looking indoor images remains a challenge due to their complex structures and diverse visual features.

Figure~\ref{fig:cate_fr_all} presents trends in SSIM, FID, CLIP-FID, and Inception Score. The `accessory' category consistently shows higher FID values, reflecting poorer content distribution, while the `animal' category achieves higher SSIM scores, demonstrating better structure-level similarity. These findings suggest that `accessory' and `indoor' images pose greater challenges for generative models in maintaining semantic consistency and achieving balanced content generation compared to other categories.

Overall, these results indicate that AI-generated images still lag behind natural images in achieving comparable levels of naturalness and image quality. Moreover, the substantial performance discrepancies across categories highlight an imbalance in the training data used for generative models. This imbalance likely limits their ability to generalize effectively across diverse content types, underscoring the need for more balanced datasets and targeted advancements in model design.

\section{Discussion}
\label{sec_app:disscussion}

\begin{table}[]
\small
\caption{Prompts trigger the safe-checker of different models.}
\label{tab:unsafe}
\renewcommand{\arraystretch}{1.2}
\centering
\begin{tabular}{c|ccc|ccc}
\hline
Models & T2I & I2I & TI2I & T2I(\textperthousand)  & I2I(\textperthousand)  & TI2I(\textperthousand)  \\ \hline
SD14 & 104 & 263 & 141  & 4.16    & 10.51   & 5.64     \\ 
SD15 & 133 & 270 & 145  & 5.32    & 10.79   & 5.80      \\ 
D·E 2 & 48 & - & -  & 9.60    & -   & -      \\ 
D·E 3 & 45 & - & -  & 9.00    & -   & -      \\ \hline
\end{tabular}

\end{table}

\paragraph{Unexpected Safety Mechanism Activation.}
In the rapidly evolving field of AI-generated content (AIGC), the balance between model robustness and safety mechanisms remains a critical area of study. A noteworthy issue has emerged with popular image synthesis models, such as Stable Diffusion v1.4, v1.5, DALL·E 2 and DALL·E 3, which occasionally fail to generate images in response to seemingly benign prompts. Some examples of these prompts are shown in Figure~\ref{fig:unsafe}. This unexpected behavior suggests an oversensitivity in the models' safety filters, which are intended to prevent the creation of inappropriate or sensitive content.

This case study delves into instances where typical prompts, which ostensibly do not contain objectionable content, nonetheless trigger these safety mechanisms. Through a series of experiments, we systematically presented a variety of benign prompts to both model versions and documented the conditions under which it is hard to output the AIGC images (Table \ref{tab:unsafe}). Our findings suggest that certain benign text, benign image, or combinations of text and image are misinterpreted by the models' safety algorithms as being potentially harmful or sensitive.

The implications of these findings are twofold. First, they highlight a critical need for refining the sensitivity of safety algorithms to reduce false positives, thereby ensuring that the AIGC models do not unduly limit creative expression or practical application. Second, they serve as a stark reminder of the challenges inherent in balancing content safety with the functional robustness of generative models. This case study aims to contribute to the ongoing discussion on optimizing safety mechanisms in AIGC models, striving to enhance their utility and accessibility without compromising essential safeguards.

\end{document}